\documentclass{article}

\PassOptionsToPackage{numbers, compress}{natbib}

     \usepackage[preprint]{neurips_2024}

\usepackage[utf8]{inputenc} 
\usepackage[T1]{fontenc}    
\usepackage{hyperref}       
\usepackage{url}            
\usepackage{booktabs}       
\usepackage{amsfonts}       
\usepackage{nicefrac}       
\usepackage{microtype}      
\usepackage{xcolor}         
\usepackage{graphicx}
\usepackage{subcaption}
\usepackage{tabularx}
\usepackage{verbatim}
 \usepackage{multirow} 
\usepackage{enumitem}
\setlist[itemize]{leftmargin=0.2cm}
\usepackage{comment}

\title{Structure-based Drug Design Benchmark: Do 3D Methods Really Dominate? }

\author{%
  Kangyu Zheng \\
 Computer Science Department \\
  Rensselaer Polytechnic Institute \\
  \texttt{zhengk5@rpi.edu} \\
  \And
  Yingzhou Lu \\
  Stanford Medicine School \\
  \texttt{lyz66@stanford.edu} \\
  \And
  Zaixi Zhang \\
  Department of Biomedical Informatics, Harvard Medical School \\
  Department of Computer Science, 
  University of Science and Technology of China (USTC) \\
  \texttt{zaixi\_zhang@hms.harvard.edu} \\
  \And 
  Zhongwei Wan \\
  Department of Computer Science and Engineering \\
  The Ohio State University \\
  \texttt{wan.512@osu.edu} \\  
  \And
  Yao Ma \\
 Computer Science Department\\
  Rensselaer Polytechnic Institute \\
  \texttt{may13@rpi.edu} \\
  \And
  Marinka Zitnik \\
  Department of Biomedical Informatics \\
  Harvard Medical School \\
  \texttt{marinka@hms.harvard.edu} \\
  \And
  Tianfan Fu \\
  Computer Science Department\\
  Rensselaer Polytechnic Institute \\
  \texttt{fut2@rpi.edu} \\  
}

\begin{document}

\maketitle

\begin{abstract}
Currently, the field of structure-based drug design is dominated by three main types of algorithms: search-based algorithms, deep generative models, and reinforcement learning. While existing works have typically focused on comparing models within a single algorithmic category, cross-algorithm comparisons remain scarce. In this paper, to fill the gap, we establish a benchmark to evaluate the performance of sixteen models across these different algorithmic foundations by assessing the pharmaceutical properties of the generated molecules and their docking affinities with specified target proteins. We highlight the unique advantages of each algorithmic approach and offer recommendations for the design of future SBDD models. 
We emphasize that 1D/2D ligand-centric drug design methods can be used in SBDD by treating the docking function as a black-box oracle, which is typically neglected. 
The empirical results show that 1D/2D methods achieve competitive performance compared with 3D-based methods that use the 3D structure of the target protein explicitly. 
Also, AutoGrow4, a 2D molecular graph-based genetic algorithm, dominates SBDD in terms of optimization ability. 
The relevant code is available in \url{https://github.com/zkysfls/2024-sbdd-benchmark}. 
\end{abstract}

\section{Introduction}

Novel types of safe and effective drugs are needed to meet the medical needs of billions worldwide and improve the quality of human life. 
The process of discovering a new drug candidate and developing it into an approved drug for clinical use is known as \textit{drug discovery}~\citep{sinha2018drug}. 
This complex process is fundamental to the development of new therapies that can manage, cure, or alleviate the symptoms of various health conditions.

Structure-based drug design (SBDD)~\cite{bohacek1996art} represents a core strategy within the drug discovery process, which utilizes the three-dimensional (3D) structures of proteins associated with diseases to develop drug candidates, serving as a fundamental method to expedite the drug discovery process through physical simulation and data-driven modeling. Based on the lock and key model~\cite{tripathi2017molecular}, molecules that bind more effectively to a disease target tend to inhibit their abnormal activity or modulate their function in a way that contributes to disease treatment, a phenomenon that has been confirmed through experimental studies~\cite{honarparvar2014integrated,blundell1996structure,lu2022cot}.

Currently, three main algorithmic approaches dominate the drug design field~\citep{Brown_2019,gao2022sample,du2022molgensurvey}: search-based algorithms like genetic algorithms (GA)~\citep{jensen2019graph,Spiegel2020,tripp2023genetic,fu2022reinforced}, deep generative models (a.k.a. generative model) like variational autoencoder (VAE)~\citep{G_mez_Bombarelli_2018} and autoregressive models~\citep{luo2021sbdd,peng2022pocket2mol,Zhang2023}, and reinforcement learning (RL) models~\citep{denovo,moldqn}. Also, there is a trend that represents the target protein in 3D format~\citep{Zhang2023,luo2021sbdd,fu2022reinforced,peng2022pocket2mol}. These models are often regarded as state-of-the-art due to the high validity, diversity, and synthesizability of their generated molecules. However, comparisons among these models remain unclear for several reasons. Firstly, current benchmarks or survey papers tend to compare models within the same algorithmic category, with a particular focus on deep generative models~\citep{du2022molgensurvey}. Secondly, most existing benchmarks emphasize the properties of the molecules themselves, neglecting the evaluation of protein-ligand interactions, which are crucial for real-world applications~\citep{Brown_2019,gao2022sample}.

To fill this blank, this paper curates a comprehensive benchmark that encompasses sixteen models spanning all three algorithmic approaches. We assess their generated molecules not only through typical heuristic molecular property oracles but also by evaluating docking scores that reflect the quality of interactions between molecules and target proteins (associated to disease). Our analysis of the top-1/10/50/100 scores from each oracle reveals that search-based algorithms, particularly genetic algorithms, generally outperform others. Also, explicit utilization of 3D structure of the target protein has not shown significant improvement compared to 2D methods. While there are some drawbacks in certain aspects, these could potentially be mitigated by integrating other algorithmic strategies.

\section{Related Works}
There has been significant progress in benchmarking efforts for drug design evaluation~\citep{Brown_2019,tripp2021fresh,huang2021therapeutics,gao2022sample,10.3389/fphar.2020.565644,harris2023benchmarking}. Specifically, Guacamol~\citep{Brown_2019} encompasses five molecule design algorithms, develops twenty novel objective functions, and assesses their performance comprehensively. Molecular Sets (MOSES)~\citep{10.3389/fphar.2020.565644} concentrates on five generative-based models (recurrent neural network (RNN), Adversarial Auto-Encoder (AAE)~\citep{makhzani2015adversarial}, Variational Auto-Encoder (VAE)~\citep{G_mez_Bombarelli_2018}), introducing eleven oracles that primarily evaluate the novelty and uniqueness of the generated molecules. Practical Molecule Optimization (PMO)~\citep{gao2022sample} offers a benchmark for twenty-five molecule design models across twenty-three objectives, providing a broad evaluation landscape. POSECHECK~\citep{harris2023benchmarking} assesses five generative-based models by employing four physical oracles to gauge the quality of protein-ligand interactions, contributing to the nuanced understanding of model efficacy in simulating realistic biochemical interactions. Recently, \citep{tripp2023genetic} designed a simple genetic algorithm on molecules based on~\citep{jensen2019graph} and compared it with several other molecule generation algorithms. The results show that genetic algorithms perform at least as well as many more complicated methods in the unconditional molecule generation task.

\paragraph{1D Molecule Design Methods}
1D molecule design methods use Simplified Molecular-Input Line-Entry System (SMILES)~\citep{doi:10.1021/ci00057a005} or SELF-referencing Embedded Strings (SELFIES)~\citep{Krenn_2020} strings as the representation of molecules. Most 1D methods produce molecule strings in an autoregressive manner. In this paper, we discuss several methods that were developed to produce molecule strings, either SMILES or SELFIES strings, including REINVENT~\citep{denovo}, SMILES and SELFIES VAE~\citep{G_mez_Bombarelli_2018}, SMILES GA~\citep{10.1246/cl.180665}, SMILES-LSTM-HC~\citep{Brown_2019}, and Pasithea~\citep{Shen_2021}. Although SELFIES string has the advantage of enforcing chemical validity rules compared to SMILES, through thorough empirical studies, \cite{gao2022sample} showed that SELFIES string-based methods do not demonstrate superiority over SMILES string-based ones.

\paragraph{2D Molecule Design Methods}
Compared to 1D molecule design methods, representing molecules using 2D molecular graphs is a more sophisticated approach. molecular 2D representation, graphs are used to depict molecules, where edges represent chemical bonds and nodes represent atoms. There are two main strategies for constructing these graphs: atom-based and fragment-based. Atom-based methods operate on one atom or bond at a time, searching the entire chemical space. On the other hand, fragment-based methods summarize common molecular fragments and operate on one fragment at a time, which can be more efficient. In this paper, we discuss several methods belonging to this category: MolDQN~\citep{moldqn}, which uses an atom-based strategy, and Graph GA~\citep{article}, Multi-constraint Molecule Sampling (MIMOSA)~\citep{fu2021mimosa}, Differentiable Scaffolding Tree (DST)~\citep{fu2022differentiable}, and AutoGrow4~\citep{Spiegel2020}, which use fragment-based strategies.

\paragraph{3D Molecule Design Methods}
Both 1D and 2D molecule design methods are ligand-centric, focusing primarily on designing the molecule itself. In structure-based drug design, as pointed out in~\citep{huang2021therapeutics}, these models take the docking function as a black box, which inputs a molecule and outputs the binding affinity score. However, these models fail to incorporate target protein structure information and consequently suffer from high computational time (to find binding pose). In contrast, 3D structure-based drug design methods take the three-dimensional geometry of the target protein as input and directly generate pocket-aware molecules in the pocket of target protein. In this paper, we cover four cutting-edge structure-based drug design methods: PocketFlow~\citep{article}, 3DSBDD~\citep{luo2021sbdd}, Pocket2mol~\citep{peng2022pocket2mol}, and ResGen~\citep{Zhang2023}.



\begin{table*}[t]
\caption{Representative structure-based drug design methods, categorized based on the molecular assembly strategies and the optimization algorithms. Columns are various molecular assembly strategies while rows are different optimization algorithms. } 
\begin{center}
\begin{small}
\resizebox{\textwidth}{!}{
\begin{tabular}{cp{4cm}p{4cm}p{4cm}c}
\toprule[1.2pt]
 & 1D SMILES/SELFIES & 2D Molecular Graph & 3D Structure-based  \\ 
 \midrule
\multirow{1}{*}{Genetic Algorithm (GA)} & SMILES-GA~\citep{10.1246/cl.180665} & AutoGrow4~\citep{Spiegel2020}, graph GA~\citep{jensen2019graph} & - \\ 
\midrule
\multirow{1}{*}{Hill Climbing} & SMILES-LSMT-HC~\citep{Brown_2019} & \multirow{1}{*}{MIMOSA~\citep{fu2021mimosa}} & - \\ \midrule
\multirow{1}{*}{Reinforcement Learning (RL)} & REINVENT~\citep{denovo} & \multirow{1}{*}{MolDQN~\citep{moldqn}} & - \\ 
\midrule 
 Gradient Ascent (GRAD) & Pasithea~\citep{Shen_2021} & DST~\citep{fu2022differentiable} & \\ 
\midrule 
\multirow{2}{*}{Generative Models} & SMILES/SELFIES-VAE-BO~\citep{G_mez_Bombarelli_2018} & - & 3DSBDD\citep{luo2021sbdd}, Pocket2mol\citep{peng2022pocket2mol}, PocketFlow\citep{article}, ResGen\citep{Zhang2023} \\   
\bottomrule[1.2pt]
\end{tabular}
}
\end{small}
\end{center}
\label{tab:methods}
\end{table*}

\section{Models}

In this paper, the models we select for evaluation are based on one or a combination of the following algorithms. For ease of comparison, we categorize all the methods based on optimization algorithm and molecular assembly strategy in Table~\ref{tab:methods}.

\textbf{Screening}: Screening (high-throughput screening) is a traditional drug design approach that searches over a library of molecules. However, it is only able to search the known drug molecular space, but is not able to explore unknown chemical space and identify novel/unknown molecules. The known chemical space ($<10^{15}$) is only taking a tiny fraction of the whole drug-like molecular space (around $10^{60}$)~\citep{bohacek1996art}. In our evaluation, we use screening as a baseline method, which randomly searches ZINC 250k library~\citep{doi:10.1021/ci3001277}.\\
\textbf{Genetic Algorithm (GA)}: Inspired by natural selection, genetic algorithm is a combinatorial optimization method that evolves solutions to problems over many generations. Specifically, in each generation, GA will perform crossover and mutation over a set of candidates to produce a pool of offspring and keep the top-k offspring for the next generation, imitating the natural selection process. In our evaluation, we choose three GA models: SMILES GA~\cite{10.1246/cl.180665} that performs GA over SMILES string-based space, Graph GA~\citep{article} that searches over atom- and fragment-level by designing their crossover and mutation rules on graph matching and AutoGrow4~\citep{Spiegel2020} which introduce another procedure called elitism that filtering the candidates by pre-defined rules. \\ 
\textbf{Variational Auto-Encoder (VAE)}: The aim of variational autoencoder is to generate new data that is similar to training data. In the molecule generation area, VAE learns a bidirectional map between molecule space and continuous latent space and optimizes the latent space. VAE itself generated diverse molecules that are learned from the training set. 
After training VAE, Bayesian optimization (BO) is used to navigate latent space efficiently, identify desirable molecules, and conduct molecule optimization. 
In our evaluation, we select two VAE-based models: SMILES-VAE-BO~\citep{G_mez_Bombarelli_2018} uses SMILES string as the input to the VAE model, and SELFIES-VAE-BO uses the same algorithm but uses SELFIES string as the molecular representation. \\
\textbf{Auto-regressive}: An auto-regressive model is a type of statistical model that is based on the idea that past values in the series can be used to predict future values. In molecule generation, an auto-regressive model would typically take the generated atom sequence as input and predict which atom would be the next. In our evaluation, we choose four auto-regressive models: PocketFlow~\citep{article} is an autoregressive flow-based generative model. 3DSBDD~\citep{luo2021sbdd} based on conventional Markov Chain Monte Carlo (MCMC) algorithms and Pocket2mol~\citep{peng2022pocket2mol} choose graph neural networks (GNN) as the backbone. Inspired by Pocket2mol, ResGen~\citep{Zhang2023} used a hierarchical autoregression, which consists of a global autoregression for learning protein-ligand interactions and atomic component autoregression for learning each atom’s topology and geometry distributions. Also, note that these models use 3D representations of target proteins.\\
\textbf{Hill Climbing (HC)}: Hill Climbing (HC) is an optimization algorithm that belongs to the family of local search techniques~\citep{selman2006hill}. It is used to find the best solution to a problem among a set of possible solutions. In molecular design, Hill Climbing would tune the generative model with the reference of generated high-scored molecules. In our evaluation, we adopt two HC models: SMILES-LSTM-HC~\citep{Brown_2019} uses an LSTM model to generate molecules and uses the HC technique to fine-tune it. MultI-constraint MOlecule SAmpling (MIMOSA)~\citep{fu2021mimosa} uses a graph neural network instead and incorporates it with HC. \\
\textbf{Gradient Ascent (GRAD)}: Similar to gradient descent, gradient ascent also estimates the gradient direction but chooses the maximum direction. In molecular design, the GRAD method is often used in molecular property function to optimize molecular generation. In our evaluation, we choose two GRAD-based models: Pasithea~\citep{Shen_2021} uses SELFIES as input and applies GRAD on an MLP-based molecular property prediction model. Differentiable Scaffolding Tree (DST)~\citep{fu2022differentiable} uses differentiable molecular graph as input and uses a graph neural network to estimate objective and the corresponding gradient. \\ 
\textbf{Reinforcement Learning (RL)}: In molecular generation context, a reinforcement learning model would take a partially-generated molecule (either sequence or molecular graph) as state; action is how to add a token or atom to the sequence or molecular graph respectively; and reward is the property score of current molecular sequence. In our evaluation, we test on two RL-based models: REINVENT~\citep{denovo} is a policy-gradient method that uses RNN to generate molecules and MolDQN~\citep{moldqn} uses a deep Q-network to generate molecular graph. \\ 
All the methods used in this paper are summarized in Table~\ref{tab:methods} for ease of comparison. 

\section{Experiments} 
\label{sec:experiment}
In this section, we demonstrate the experimental results. We start with the description of experimental setup. Then, we present and analyze the experimental results, including protein-ligand bindings, pharmaceutical properties of generated molecules (e.g., drug-likeness and synthetic accessibility), and other qualities of generated molecules (e.g., diversity, validity).

\subsection{Experimental Setup}
\label{sec:experimental_setup}
\subsubsection{Oracle} 
\label{sec:oracle}
In drug discovery, we need to evaluate the pharmaceutical properties of the generated molecules, such as binding affinity to certain target proteins, drug-likeness, synthetic accessibility, solubility, etc. These property evaluators are also known as \textit{oracle}. 
In this section, we introduce the oracle we chose to evaluate these models. All our oracle functions come from Therapeutic Data Commons (TDC)~\citep{Huang2022artificial,huang2021therapeutics}\footnote{\url{https://tdcommons.ai/functions/oracles/}}. 

\noindent\textbf{(1) Docking Score}: Molecular docking is a measurement of free energy exchange between a ligand and a target protein during the binding process. A lower docking score means the ligand would have a higher potential to pose higher bioactivity with a given target. Compared with other heuristic oracles, such as QED (quantitative estimate of drug-likeness), and LogP (Octanol-water partition coefficient), docking reflects the binding affinities between drug molecule and target \citep{Graff_2021}. Our experiments use \textbf{TDC.Docking} oracle function, which is based on AutoDock Vina~\citep{doi:10.1021/acs.jcim.1c00203} to test with these models.
We chose seven representative and diverse target proteins in the TDC docking benchmark, which are selected from CrossDock~\citep{francoeur2020three}. 
The PDBIDs are 1iep, 3eml, 3ny8, 4rlu, 4unn, 5mo4, 7l11. These crystallography structures are across different fields, including virology, immunology, and oncology~\citep{Huang2022artificial,huang2021therapeutics,lu2019integrated}. They cover various kinds of diseases such as chronic myelogenous leukemia, tuberculosis, SARS-COVID-2, etc. They represent a breadth of functionality, from viral replication mechanisms to cellular signaling pathways and immune responses. \\ 
\noindent\textbf{(2) Heuristic Oracles}: Although heuristic oracles are considered to be ``trivial'' and too easily optimized, we still incorporate some of them into our evaluation metrics for comprehensive analysis. In our experiments, we utilize Quantitative Estimate of Drug-likeness (QED), SA, and LogP as our heuristic oracles. QED evaluates a molecule's drug-likeness on a scale from 0 to 1, where 0 indicates minimal drug-likeness and 1 signifies maximum drug-likeness, aligning closely with the physicochemical properties of successful drugs. SA, or Synthetic Accessibility, assesses the ease of synthesizing a molecule, with scores ranging from 1 to 10; a lower score suggests easier synthesis. LogP measures a compound's preference for a lipophilic (oil-like) phase over a hydrophilic (water-like) phase, essentially indicating its solubility in water, where the optimal range depends on the type of drug. But mostly the value should be between 0 and 5~\citep{Krenn_2020}.\\
\noindent\textbf{(3) Molecule Generation Oracles}: While docking score oracles and heuristic oracles focus on evaluating individual molecules, molecule generation oracles assess the quality of all generated molecules as a whole. In our experiments, we choose three metrics to evaluate the generated molecules of each model: diversity, validity, and uniqueness. Diversity is measured by the average pairwise Tanimoto distance between the Morgan fingerprints~\citep{benhenda2017chemgan}. Validity is determined by checking atoms' valency and the consistency of bonds in aromatic rings using RDKit's molecular structure parser~\citep{10.3389/fphar.2020.565644}. Uniqueness is measured by the frequency at which a model generates duplicated molecules, with lower values indicating more frequent duplicates~\citep{10.3389/fphar.2020.565644}.

\subsubsection{Model Setup} 
\label{sec:model_setup}
\begin{figure*}[t]
\vspace{-12pt}
\centering
\includegraphics[width=0.9\textwidth]{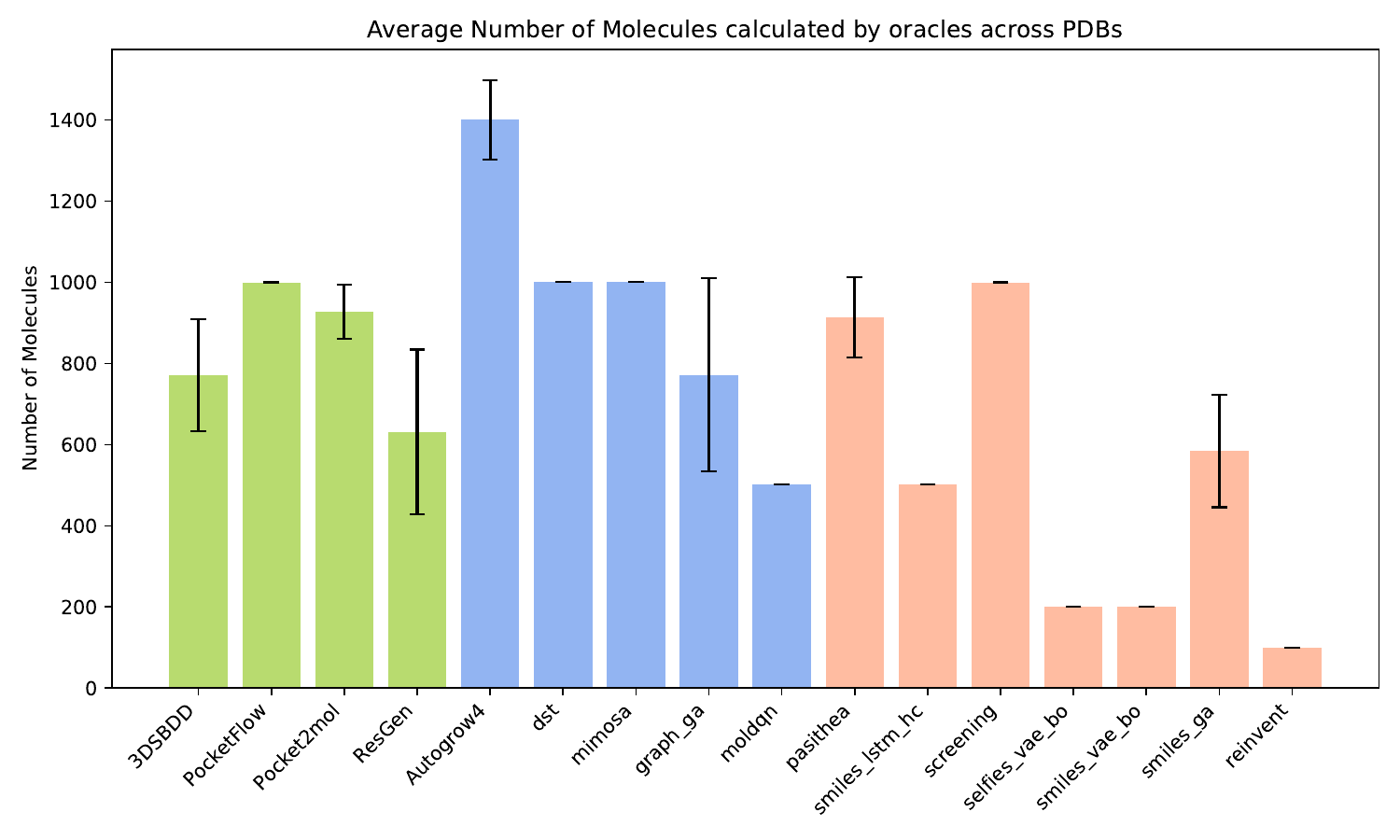}
\caption{The bar chart of average generated molecules that are calculated by our selected oracles for each model across all target proteins under given time. 1D methods are colored red, blue is used to indicate 2D methods, and green represents 3D methods.}
\label{fig:avg_gen_bar}
\vspace{-12pt}
\end{figure*}

For each model, we generate 1,000 molecules for each given target protein, and each molecule is evaluated by the TDC oracle functions~\citep{huang2021therapeutics,Huang2022artificial}. Each experiment is run on one OSC Ascend node~\citep{OhioSupercomputerCenter1987} for 96 hours, which is the maximum time allowed for a single experiment, and we only run each model once. Five models (3DSBDD, PocketFlow, Pocket2mol, ResGen, and AutoGrow4) first generate a certain number of molecules within the given time, and then we run oracle functions on each molecule. All the other models come from the PMO benchmark~\citep{gao2022sample}, and our experiment follows its setting, where each molecule is first generated, then the oracle function is used to calculate the score, and then the model moves on to generate the second molecule. None of the tested models have prior knowledge of these oracle functions. Among all the models, five of them manage to generate and have been evaluated by the oracle for 1,000 or more molecules within the given time across all target proteins (AutoGrow4, PocketFlow, DST, MIMOSA, and Screening); other models do not generate enough molecules or have not been evaluated for enough molecules within the given 96 hours, mostly because the docking oracle function is time-consuming. The VAE models and REINVENT only generate 200 and 100 molecules, respectively, because we observe that when they generate more than this number of molecules, the models crash. Figure~\ref{fig:avg_gen_bar} shows the average number of molecules that has been generated and successfully evaluated by oracle functions within experiment time.

\begin{table}[t]

\caption{The average of each model's Top-10 Docking score for each target protein. } 
\label{top10-docking-table}
\vskip 0.15in
\begin{center}
\begin{small}
\begin{sc}

\begin{tabular}{cccccccccccccc}
\toprule
Model \ & 1iep & 3eml & 3ny8 & 4rlu \\
\midrule
3DSBDD & -9.05 $\pm$ 0.38 & -10.02 $\pm$ 0.15 & -10.10 $\pm$ 0.24 & -9.80 $\pm$ 0.55 \\
AutoGrow4 & \textbf{-13.23} $\pm$ 0.11 & \textbf{-13.03} $\pm$ 0.09 & -11.70 $\pm$ 0.00 & -11.20 $\pm$ 0.00 \\
Pocket2mol & -10.17 $\pm$ 0.53 & -12.25 $\pm$ 0.27 & \textbf{-11.89} $\pm$ 0.16 & -10.57 $\pm$ 0.12 \\
PocketFlow & -12.49 $\pm$ 0.70 & -9.25 $\pm$ 0.29 & -8.56 $\pm$ 0.35 & -9.65 $\pm$ 0.25 \\
ResGen & -10.97 $\pm$ 0.29 & -9.25 $\pm$ 0.95 & -10.96 $\pm$ 0.42 & \textbf{-11.75} $\pm$ 0.42 \\
DST & -10.95 $\pm$ 0.57 & -10.67 $\pm$ 0.24 & -10.54 $\pm$ 0.22 & -10.88 $\pm$ 0.37 \\
graph GA & -10.03 $\pm$ 0.41 & -9.89 $\pm$ 0.25 & -9.94 $\pm$ 0.15 & -10.22 $\pm$ 0.39 \\
MIMOSA & -10.96 $\pm$ 0.57 & -10.69 $\pm$ 0.24 & -10.51 $\pm$ 0.23 & -10.81 $\pm$ 0.39 \\
MolDQN & -6.73 $\pm$ 0.12 & -6.51 $\pm$ 0.15 & -7.09 $\pm$ 0.16 & -6.79 $\pm$ 0.26 \\
Pasithea & -10.86 $\pm$ 0.29 & -10.31 $\pm$ 0.09 & -10.69 $\pm$ 0.27 & -10.92 $\pm$ 0.35 \\
reinvent & -9.87 $\pm$ 0.31 & -9.48 $\pm$ 0.39 & -9.61 $\pm$ 0.36 & -9.69 $\pm$ 0.29 \\
screening & -10.86 $\pm$ 0.26 & -10.90 $\pm$ 0.54 & -10.73 $\pm$ 0.45 & -10.86 $\pm$ 0.22 \\
selfies-vae-bo & -10.15 $\pm$ 0.60 & -9.76 $\pm$ 0.12 & -9.99 $\pm$ 0.28 & -10.00 $\pm$ 0.23 \\
smiles ga & -9.56 $\pm$ 0.17 & -9.56 $\pm$ 0.37 & -10.00 $\pm$ 0.26 & -9.61 $\pm$ 0.19 \\
smiles lstm hc & -10.38 $\pm$ 0.21 & -10.30 $\pm$ 0.15 & -10.19 $\pm$ 0.12 & -10.49 $\pm$ 0.49 \\
SMILES-VAE-BO & -9.93 $\pm$ 0.22 & -9.78 $\pm$ 0.10 & -9.96 $\pm$ 0.29 & -10.05 $\pm$ 0.20 \\
\bottomrule
\end{tabular}

\end{sc}
\end{small}
\end{center}

\begin{center}
\begin{small}
\begin{sc}

\begin{tabular}{cccccccccccccc}
\toprule
Model \ & 4unn & 5mo4 & 7l11 \\
\midrule
3DSBDD & -8.23 $\pm$ 0.30 & -8.71 $\pm$ 0.45 & -8.47 $\pm$ 0.18 \\
AutoGrow4 & -11.14 $\pm$ 0.12 & \textbf{-10.38} $\pm$ 0.27 & -8.84 $\pm$ 0.33 \\
Pocket2mol & \textbf{-12.20} $\pm$ 0.34 & -10.07 $\pm$ 0.62 & \textbf{-9.74} $\pm$ 0.38 \\
PocketFlow & -7.90 $\pm$ 0.78 & -7.80 $\pm$ 0.42 & -8.35 $\pm$ 0.31 \\
ResGen & -9.41 $\pm$ 0.23 & -10.34 $\pm$ 0.39 & -8.74 $\pm$ 0.24 \\
DST & -9.71 $\pm$ 0.19 & -10.03 $\pm$ 0.36 & -8.33 $\pm$ 0.41 \\
graph GA & -9.32 $\pm$ 0.51 & -9.29 $\pm$ 0.20 & -7.75 $\pm$ 0.32 \\
MIMOSA & -9.66 $\pm$ 0.25 & -10.02 $\pm$ 0.36 & -8.33 $\pm$ 0.41 \\
molDQN & -5.92 $\pm$ 0.26 & -6.27 $\pm$ 0.10 & -6.87 $\pm$ 0.20 \\
Pasithea & -9.69 $\pm$ 0.32 & -9.77 $\pm$ 0.21 & -8.06 $\pm$ 0.22 \\
REINVENT & -8.70 $\pm$ 0.25 & -8.92 $\pm$ 0.38 & -7.25 $\pm$ 0.21 \\
screening & -9.80 $\pm$ 0.23 & -9.91 $\pm$ 0.30 & -8.15 $\pm$ 0.26 \\
SELFIES VAE BO & -9.02 $\pm$ 0.33 & -9.18 $\pm$ 0.39 & -7.75 $\pm$ 0.22 \\
SMILES GA & -8.80 $\pm$ 0.20 & -9.21 $\pm$ 0.23 & -7.54 $\pm$ 0.32 \\
SMILES LSTM HC & -9.36 $\pm$ 0.17 & -9.71 $\pm$ 0.43 & -7.90 $\pm$ 0.26 \\
SMILES-VAE-BO & -9.03 $\pm$ 0.30 & -9.18 $\pm$ 0.39 & -7.74 $\pm$ 0.25 \\
\bottomrule
\end{tabular}

\end{sc}
\end{small}
\end{center}
\end{table}

\begin{figure*}[t!]
\centering
\includegraphics[width=0.99\textwidth]{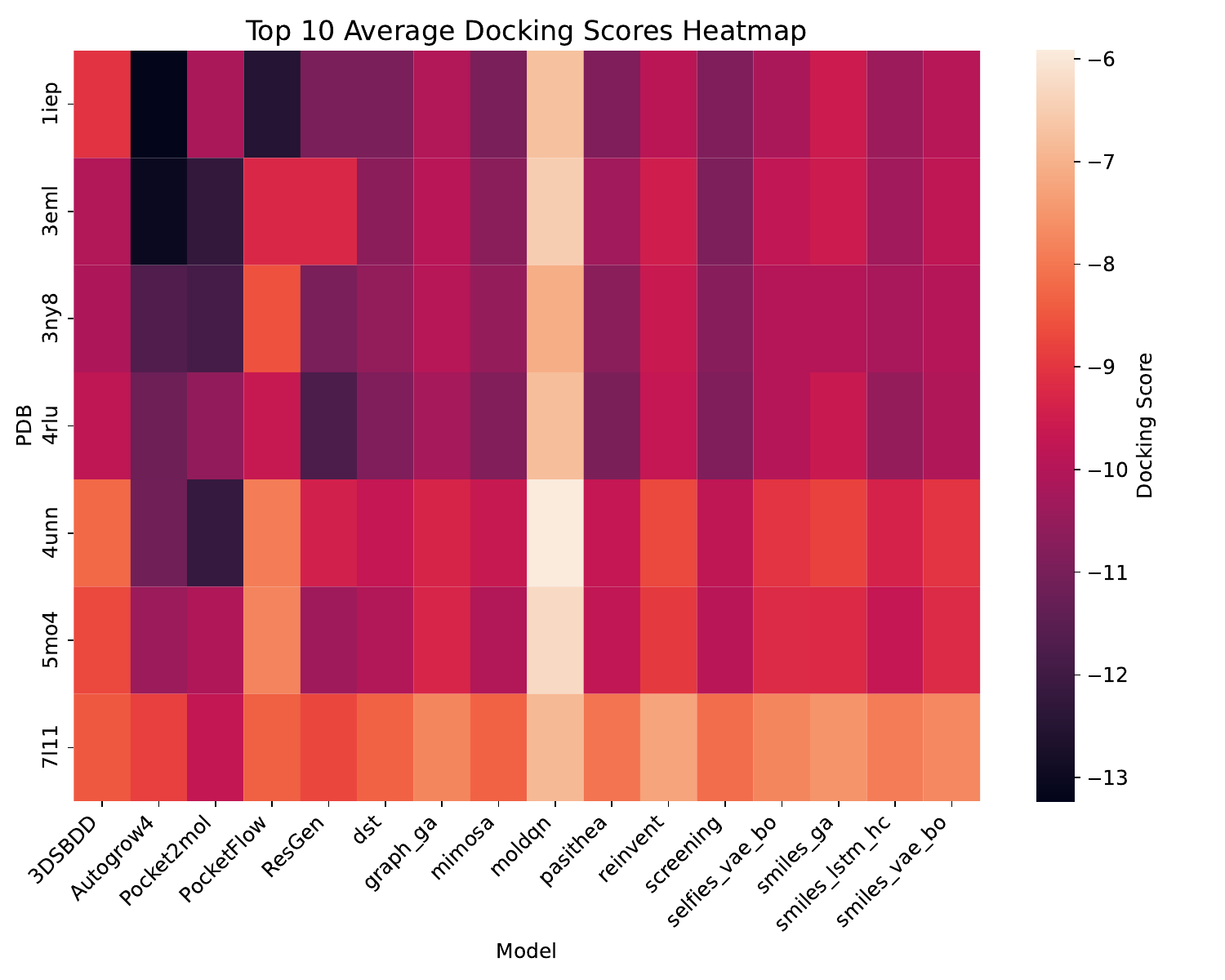}
\caption{The heatmap based on the average of each model's Top-10 docking score for each target protein.}
\label{fig:top_10_docking_heatmap}

\end{figure*}

\begin{figure*}[t!]

 \centering
    \hspace{0cm}
    \begin{subfigure}{0.49\textwidth}
        \centering
        \includegraphics[width=\textwidth]{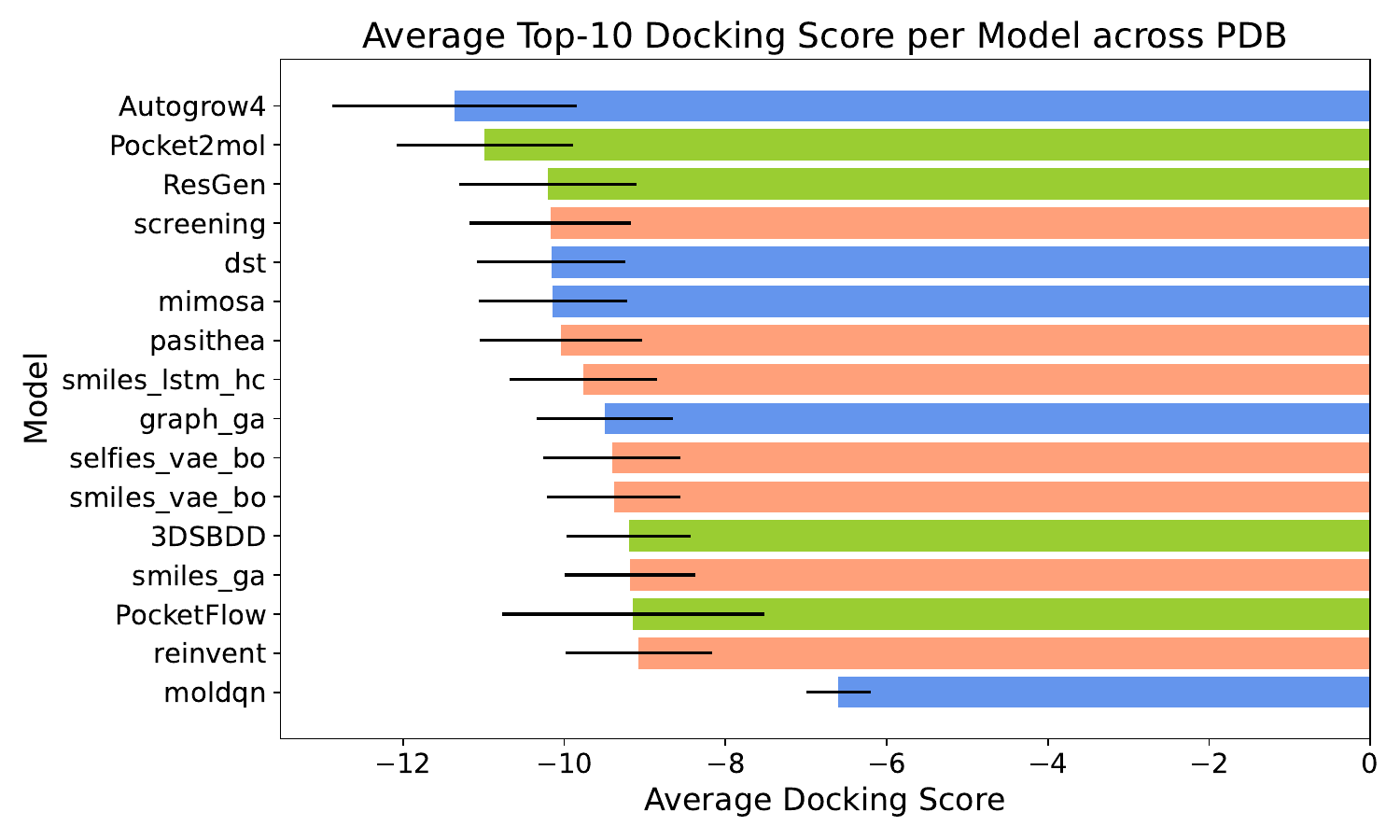}
        \label{fig:bar_chart_bydim}
    \end{subfigure}
    \hspace{0cm}
    \begin{subfigure}{0.49\textwidth}
        \centering
        \includegraphics[width=\textwidth]{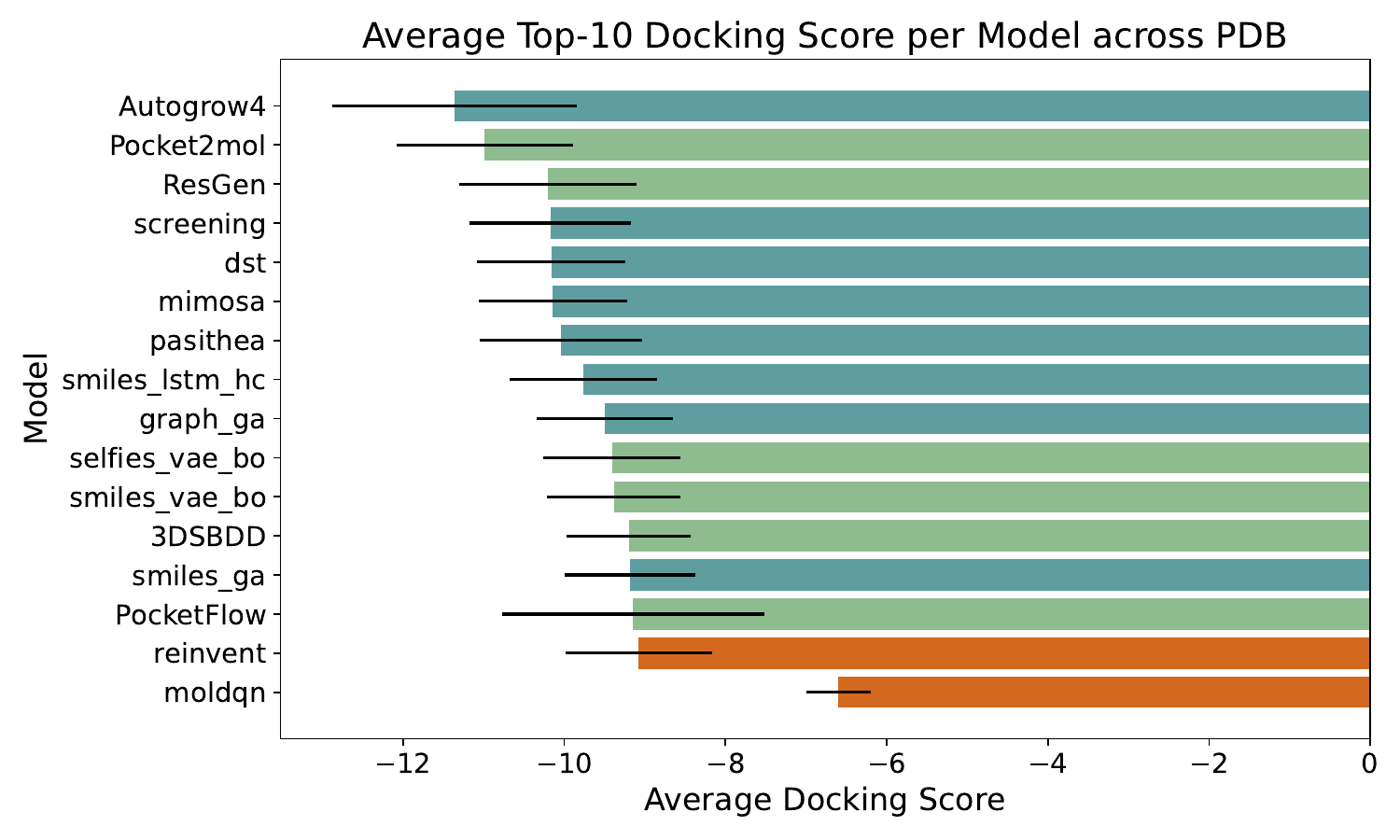}
        \label{fig:bar_chart_bytype}
    \end{subfigure}
    \hspace{0cm}
\caption{The bar chart is based on the average of each model's Top-10 docking score. On the left, 1D methods are colored red, blue is used to indicate 2D methods, and green represents 3D methods. On the right, generative models are in blue, search-based methods are in green, and reinforcement learning methods are in red.
}
\label{fig:bar_chart_avg}

\end{figure*}


\subsection{Experimental Results}
\subsubsection{Results of Binding Affinities}
\noindent\textbf{Overall Performance}: Overall, search-based algorithms (including Screening, Genetic Algorithm [GA], Hill Climbing [HC], and Gradient-based methods [GRAD]) demonstrate superior performance compared to generative models (like VAE and Auto-regressive) and reinforcement learning-based algorithms. Although generative models have good performance on Top-1 docking score (Appendix~\ref{sec: additional_table_fig} Table~\ref{top1-docking-table}), search-based algorithms take advantages on Top-10/50/100 docking score, as shown in Table~\ref{top10-docking-table} to~\ref{top100-docking-table}.\\
\noindent\textbf{Search-based Algorithms}: Among all the search-based algorithms, AutoGrow4 exhibits the best performance. This superiority is not only reflected in its consistently highest scores in the Top 1/10/50/100 categories but also in the outstanding docking scores of the majority of its generated molecules compared to other methods across every target protein, as indicated by Appendix~\ref{sec: additional_table_fig} Table~\ref{top1-docking-table} to~\ref{top100-docking-table}. We believe that the elitism procedure incorporated in AutoGrow4 enhances its performance by providing better candidates for crossover and mutation. While other search-based methods also perform well overall, no single algorithm category within this group distinctly outperforms the others. It appears that neither the format of the input (such as SMILES/SELFIES or scaffold graph) nor the specific algorithm employed has a significant impact on performance.\\ 
\noindent\textbf{Generative Models}: In our experiments, we evaluated two categories of generative models: Variational Autoencoders (VAEs) and Auto-regressive models. Firstly, two VAE-based models, SMILES-VAE-BO and SELFIES-VAE-BO, demonstrated consistent performance across all target proteins, with most of their generated molecules achieving docking scores between -6 and -9. However, only a few molecules from these models exceeded a -10 score, and neither model showed a distinct advantage over the other. \\
Regarding the autoregressive models, among the four models we evaluated, Pocket2Mol demonstrates the best overall performance. This is not only because it achieves the highest scores in Top-1, Top-10, Top-50, and Top-100 rankings but also because its average top docking score remains above -10 across all these rankings. For the remaining models, ResGen ranks second, followed by PocketFlow and 3DSBDD. We also noticed that some of our selected target pockets appeared in Crossdeck~\citep{doi:10.1021/acs.jcim.0c00411}, which is the dataset that our selected autoregressive models used for training and evaluation. However, our experimental results show that these autoregressive models do not always have advantages compared to other models. For example, in target 3EML, AutoGrow4 has the best performance across Top 1 to Top 100, while in target 4UNN, Pocket2mol has the advantage.\\
\noindent\textbf{Reinforcement Learning}: We incorporate two reinforcement learning-based models: MolDQN and REINVENT. Overall, REINVENT demonstrates superior performance compared to MolDQN. The majority of molecules generated by REINVENT have docking scores around -8, whereas those by MolDQN are mostly around -6. This leads us to suspect that the policy-gradient method might be more suitable than the deep Q-network approach for the task of molecular generation. Compared to other models, reinforcement learning models do not exhibit good performance. As shown in Figure~\ref{fig:bar_chart_avg}, these two reinforcement learning models have the lowest scores in the docking score oracle. This may suggest that the reinforcement learning algorithm may not have a strong ability to produce molecules with good docking scores.

\paragraph{Example of the Generated Molecules.} Also, we show the 3D poses of molecules that have the best docking score for each target protein in Figure~\ref{fig:pose} in Appendix~\ref{sec: additional_table_fig}. We find that the generated molecules could bind tightly to the pocket of target proteins.

\subsubsection{Results of Pharmaceutical Properties}

Then, we report and analyze the pharmaceutical properties of the generated molecules. 

\noindent\textbf{SA}: Overall, most of the models generate molecules with scores between 1 to 3. Notably, 3DSBDD, MolDQN, and REINVENT produce molecules with scores ranging above 3. Additionally, these models exhibit high variance in their scores. For instance, in the case of 3DSBDD, the lowest score observed is 1, yet it can also generate molecules scoring as high as 8 for a specific target protein. \\

\noindent\textbf{QED}: Most of the models generate molecules with scores between 0.8. However, 3DSBDD and REINVENT tend to produce molecules with scores primarily in the range of 0.6 to 0.7, while MolDQN's generated molecules hover around 0.4. Overall, nearly all models have the same level of performance except reinforcement learning-based models, which have worse performances.\\ 

\noindent\textbf{LogP}: Overall, nearly all the models tested produced the majority of their molecules within the 0 to 3 range, which is deemed suitable for a drug, with the exception of MolDQN. The molecules generated by MolDQN often have a LogP score of less than 0 across all target proteins, indicating a high solubility in water. Furthermore, generative models, particularly 3DSBDD, predominantly generate molecules with scores around 0. 

\paragraph{Additional Experimental Results} 
Furthermore, we report the numerical values of top-$K$ docking, QED, SA, and LogP scores for all the methods across different target proteins in the Appendix~\ref{sec: additional_table_fig}, as well as their relative ranking on all the metrics.  

\subsubsection{Molecule Generation Quality}
In the diversity oracle, all models score above 0.8, with one model from each algorithm category exceeding 0.9: PocketFlow from generative models, Graph GA from search-based methods, and MolDQN from reinforcement learning. This suggests that these three models are particularly effective at generating diverse sets of molecules.
In the validity oracle, all models achieve a perfect score of 1, except for 3DSBDD. Similarly, in the uniqueness oracle, all models score 1, except for 3DSBDD, AutoGrow4, and PocketFlow. It is unclear why these models have lower scores in validity and uniqueness, especially when other models from the same algorithm category perform well.
One possible explanation for 3DSBDD's low validity and uniqueness scores could be issues with its molecule generation process, such as producing invalid molecular structures or duplicates. Despite their high diversity scores, AutoGrow4 and PocketFlow's lower uniqueness scores might indicate a tendency to generate similar molecules.
Further investigation into the specific architectures and training procedures of these models could provide insights into their divergent performance. It may also be valuable to analyze the trade-offs between diversity, validity, and uniqueness in molecule generation and how different models balance these objectives.
Overall, while most models demonstrate strong performance across the three oracles, the lower validity and uniqueness scores of 3DSBDD, AutoGrow4, and PocketFlow highlight the importance of evaluating multiple aspects of generated molecules to assess model performance comprehensively.

\subsubsection{Key observations}
We summarize the following insightful observations drawn from the experimental results, which benefits design of future SBDD models.
\begin{itemize}
\item Most structure-based drug design method uses the 3D structure of the target protein explicitly and grow the drug molecules in the pocket of the target protein. We pinpoint another direction that regards the docking function as a black box and uses 1D/2D ligand-centric methods to produce drug molecules, which is usually neglected by the community. In this paper, we empirically prove that this kind of method would achieve superior performance. 
\item Generally, 3D SBDD algorithms (using 3D target protein structure explicitly) do not demonstrate significant superiority over 2D methods. 
\item No methods can dominate structure-based drug design in all the evaluation metrics (docking score, SA, QED, diversity, validity, and uniqueness), as shown in Appendix~\ref{sec: additional_table_fig} Figure~\ref{fig:radar_chart}. 
\item AutoGrow4, a 2D genetic algorithm, exhibits the best optimization performance in terms of top-K docking scores in most target proteins. Also, it owns desirable synthetic accessibility. 
\end{itemize}

\section{Conclusions}
Currently, the landscape of structure-based drug design models is vast, featuring various algorithmic backbones, yet comparative analyses across them are scarce. In this study, we design experiments to evaluate the quality of molecules generated by each model. Our experiments extend beyond conventional heuristic oracles related to molecular properties, also examining the affinity between molecules and selected target proteins. Our findings indicate that models based on genetic algorithms exhibit a higher potential for producing molecules that dock effectively with given target proteins. Also, representing target molecules in 3D format does not significantly improve both the molecular quality and blinding affinity. Although we observed that there is no single method that could excel both our two metrics,  we suggest that when developing new structure-based drug discovery models in the future, it would be advantageous to integrate genetic algorithms with other computational approaches to enhance both docking scores and molecular properties.


{\small
\bibliographystyle{unsrtnat}
\bibliography{main.bib}
}

\appendix

\section{Appendix}

\subsection{Implementation Details}
To run our selected models and evaluate their generated molecules, we establish two conda environments: Test environment and TDC environment. Test environment is used for these models to generate molecules: 3DSBDD, Autogrow4, Pocket2mol, PocketFlow and ResGen. TDC environment is used for other models that are under PMO~\citep{gao2022sample} package and evaluate all the models' generated molecules by TDC oracles. \\
For the models that are included in the PMO~\citep{gao2022sample}, we use the production mode and make a single run of producing 1000 molecules. For the rest of the models, we follow their instructions of generating molecules with PDB information. The detailed parameters and operation could be found in \url{https://github.com/zkysfls/2024-sbdd-benchmark}. 

\subsection{Limitations} 
\label{sec:limitation}
Structure-based drug design is a vast and fast-growing field, and there are important methods yet to be included in our benchmark. However, our benchmark is an ongoing effort and we strive to continuously include more state-of-the-art methods in the future. 
For example, structure-based drug design (SBDD) is utilized in the early stages of drug discovery and development. However, there is a significant gap between early-phase drug discovery and the subsequent pre-clinical and clinical drug development phases~\cite{chen2024uncertainty,fu2022hint}. This disconnect can cause drug candidates to fail during clinical trials. Therefore, incorporating feedback from the later stages of drug development to create new design criteria for SBDD may enhance therapeutic success rates.

\subsection{Additional Tables and Figures}

\label{sec: additional_table_fig}

\begin{figure*}[h]
\centering
\includegraphics[width=0.9\textwidth]{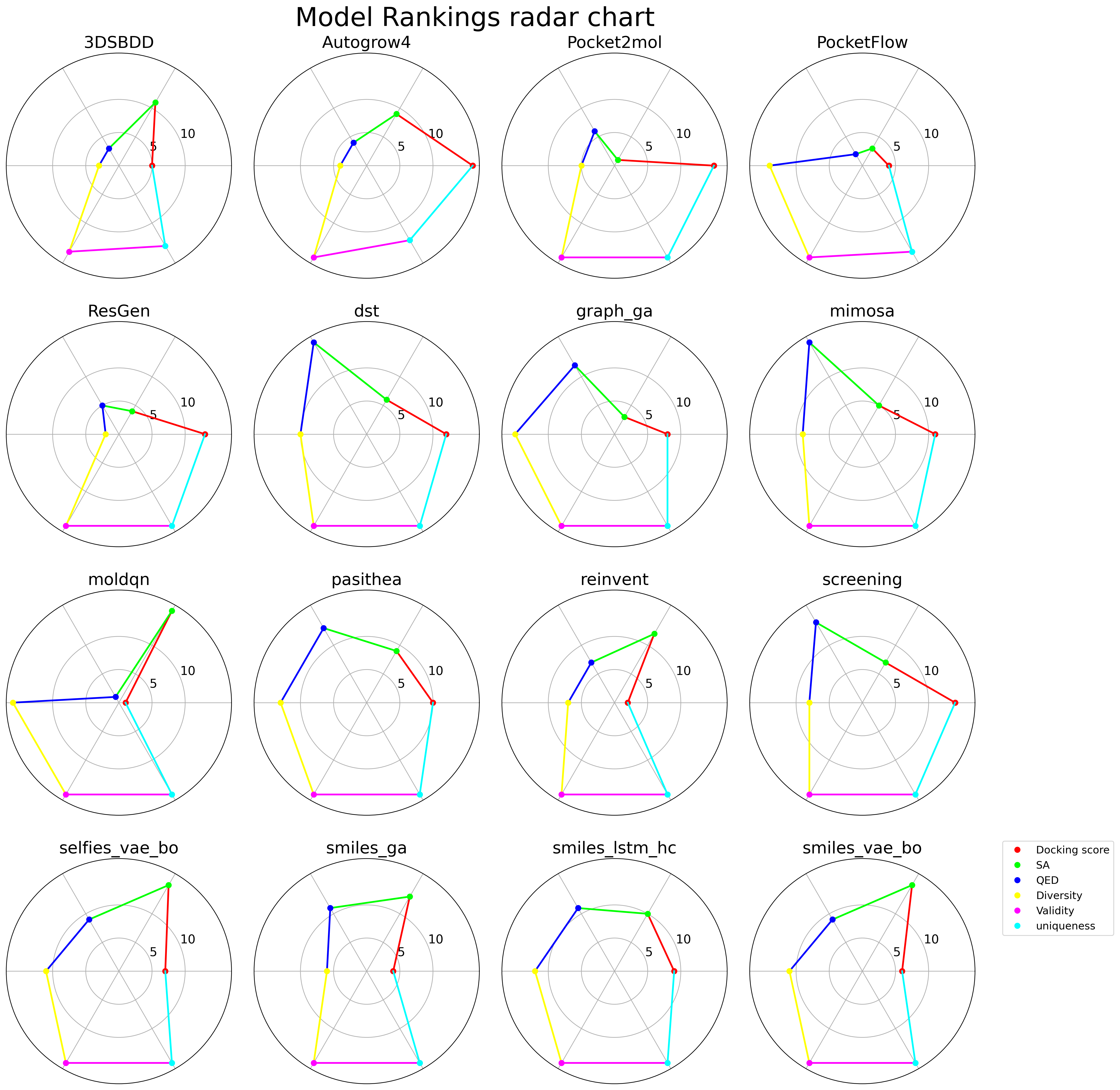}
\caption{The radar chart is based on the ranking of each model's average performance.  The outermost circle represents the best ranking for each metric and vice versa. No methods can dominate structure-based drug design in all the evaluation metrics. }
\label{fig:radar_chart}
\end{figure*}

\begin{table*}[h]
\caption{Top 1 Docking score for each target protein. } 
\label{top1-docking-table}
\vskip 0.15in
\begin{center}
\begin{small}
\begin{sc}
\begin{tabular}{cccccccccccccc}
\toprule
Model \ & 1iep & 3eml & 3ny8 & 4rlu \\
\midrule
3DSBDD & -10.00 & -10.40 & -10.60 & -11.40 \\
AutoGrow4 & -13.40 & \textbf{-13.30} & -11.70 & -11.20 \\
Pocket2mol & -11.50 & -12.80 & \textbf{-12.20} & -10.70 \\
PocketFlow & \textbf{-13.90} & -9.80 & -9.20 & -10.10 \\
ResGen & -11.60 & -10.80 & -11.70 & \textbf{-12.60} \\
DST & -12.20 & -11.00 & -11.00 & -11.40 \\
graph GA & -10.80 & -10.50 & -10.20 & -11.10 \\
MIMOSA & -12.20 & -11.10 & -11.00 & -11.40 \\
molDQN & -6.90 & -6.80 & -7.50 & -7.50 \\
Pasithea & -11.60 & -10.40 & -11.20 & -11.60 \\
REINVENT & -10.40 & -10.40 & -10.30 & -10.10 \\
screening & -11.30 & -12.20 & -11.90 & -11.20 \\
SELFIES VAE BO & -11.80 & -10.00 & -10.70 & -10.50 \\
SMILES GA & -9.90 & -10.50 & -10.60 & -9.90 \\
SMILES LSTM HC & -10.70 & -10.60 & -10.40 & -11.30 \\
SMILES-VAE-BO & -10.30 & -10.00 & -10.70 & -10.50 \\
\bottomrule
\end{tabular}
\end{sc}
\end{small}
\end{center}

\begin{center}
\begin{small}
\begin{sc}
\begin{tabular}{cccccccccccccc}
\toprule
Model \ & 4unn & 5mo4 & 7l11 \\
\midrule
3DSBDD & -8.90 & -9.80 & -8.80 \\
AutoGrow4 & -11.20 & -10.80 & -9.60 \\
Pocket2mol & \textbf{-12.80} & \textbf{-11.90} & \textbf{-10.40} \\
PocketFlow & -9.50 & -8.70 & -8.90 \\
ResGen & -9.90 & -11.00 & -9.40 \\
DST & -9.90 & -11.00 & -9.30 \\
graph GA & -10.60 & -9.60 & -8.60 \\
MIMOSA & -10.00 & -11.00 & -9.30 \\
molDQN & -6.40 & -6.50 & -7.20 \\
Pasithea & -10.40 & -10.10 & -8.60 \\
REINVENT & -9.00 & -9.60 & -7.70 \\
screening & -10.30 & -10.50 & -8.70 \\
SELFIES VAE BO & -9.60 & -10.10 & -8.20 \\
SMILES GA & -9.30 & -9.60 & -8.10 \\
SMILES LSTM HC & -9.70 & -10.60 & -8.40 \\
SMILES-VAE-BO & -9.60 & -10.10 & -8.20 \\
\bottomrule
\end{tabular}
\end{sc}
\end{small}
\end{center}
\end{table*}

\begin{table*}[h]
\caption{Top 50 Docking score for each target protein. } 
\label{top50-docking-table}
\vskip 0.15in
\begin{center}
\begin{small}
\begin{sc}
\begin{tabular}{cccccccccccccc}
\toprule
Model \ & 1iep & 3eml & 3ny8 & 4rlu \\
\midrule
3DSBDD & -8.43 $\pm$ 0.39 & -9.50 $\pm$ 0.34 & -9.15 $\pm$ 0.61 & -9.10 $\pm$ 0.47 \\
AutoGrow4 & \textbf{-12.47} $\pm$ 0.70 & \textbf{-12.45} $\pm$ 0.36 & -11.04 $\pm$ 0.38 & \textbf{-10.92} $\pm$ 0.19 \\
Pocket2mol & -9.57 $\pm$ 0.40 & -11.64 $\pm$ 0.37 & \textbf{-11.24} $\pm$ 0.42 & -10.18 $\pm$ 0.25 \\
PocketFlow & -11.60 $\pm$ 0.57 & -8.52 $\pm$ 0.47 & -7.97 $\pm$ 0.35 & -8.96 $\pm$ 0.41 \\
ResGen & -10.36 $\pm$ 0.40 & -7.22 $\pm$ 1.14 & -10.35 $\pm$ 0.39 & -10.79 $\pm$ 0.61 \\
DST & -10.07 $\pm$ 0.55 & -10.03 $\pm$ 0.37 & -10.04 $\pm$ 0.30 & -10.24 $\pm$ 0.39 \\
graph GA & -9.19 $\pm$ 0.53 & -9.19 $\pm$ 0.43 & -9.31 $\pm$ 0.40 & -9.52 $\pm$ 0.45 \\
MIMOSA & -10.09 $\pm$ 0.55 & -10.01 $\pm$ 0.39 & -10.04 $\pm$ 0.30 & -10.23 $\pm$ 0.37 \\
MolDQN & -6.31 $\pm$ 0.27 & -6.14 $\pm$ 0.24 & -6.39 $\pm$ 0.49 & -6.15 $\pm$ 0.40 \\
Pasithea & -10.10 $\pm$ 0.48 & -9.91 $\pm$ 0.25 & -10.10 $\pm$ 0.36 & -10.25 $\pm$ 0.42 \\
REINVENT & -8.53 $\pm$ 0.81 & -8.55 $\pm$ 0.58 & -8.70 $\pm$ 0.56 & -8.69 $\pm$ 0.63 \\
screening & -10.23 $\pm$ 0.40 & -10.11 $\pm$ 0.50 & -10.12 $\pm$ 0.40 & -10.24 $\pm$ 0.38 \\
SELFIES VAE BO & -9.23 $\pm$ 0.63 & -9.24 $\pm$ 0.39 & -9.35 $\pm$ 0.42 & -9.30 $\pm$ 0.43 \\
SMILES GA & -8.90 $\pm$ 0.41 & -8.65 $\pm$ 0.53 & -9.13 $\pm$ 0.52 & -9.05 $\pm$ 0.33 \\
SMILES LSTM HC & -9.61 $\pm$ 0.47 & -9.80 $\pm$ 0.33 & -9.63 $\pm$ 0.36 & -9.74 $\pm$ 0.49 \\
SMILES-VAE-BO & -9.19 $\pm$ 0.49 & -9.27 $\pm$ 0.39 & -9.34 $\pm$ 0.41 & -9.31 $\pm$ 0.45 \\

\bottomrule
\end{tabular}
\end{sc}
\end{small}
\end{center}

\begin{center}
\begin{small}
\begin{sc}
\begin{tabular}{cccccccccccccc}
\toprule
Model \ & 4unn & 5mo4 & 7l11 \\
\midrule
3DSBDD & -7.45 $\pm$ 0.52 & -7.98 $\pm$ 0.45 & -8.08 $\pm$ 0.25 \\
AutoGrow4 & -10.89 $\pm$ 0.14 & \textbf{-10.20} $\pm$ 0.16 & -8.44 $\pm$ 0.26 \\
Pocket2mol & \textbf{-11.56} $\pm$ 0.40 & -9.55 $\pm$ 0.40 & \textbf{-9.11} $\pm$ 0.40 \\
PocketFlow & -6.91 $\pm$ 0.63 & -7.06 $\pm$ 0.45 & -7.85 $\pm$ 0.31 \\
ResGen & -8.63 $\pm$ 0.50 & -9.39 $\pm$ 0.56 & -8.28 $\pm$ 0.30 \\
DST & -9.13 $\pm$ 0.34 & -9.30 $\pm$ 0.43 & -7.81 $\pm$ 0.35 \\
graph GA & -8.43 $\pm$ 0.54 & -8.66 $\pm$ 0.41 & -7.15 $\pm$ 0.38 \\
MIMOSA & -9.11 $\pm$ 0.33 & -9.32 $\pm$ 0.41 & -7.83 $\pm$ 0.35 \\
MolDQN & -5.47 $\pm$ 0.28 & -5.80 $\pm$ 0.33 & -5.98 $\pm$ 0.52 \\
Pasithea & -9.02 $\pm$ 0.41 & -9.24 $\pm$ 0.33 & -7.68 $\pm$ 0.23 \\
REINVENT & -7.89 $\pm$ 0.56 & -7.98 $\pm$ 0.58 & -6.64 $\pm$ 0.40 \\
screening & -9.17 $\pm$ 0.38 & -9.38 $\pm$ 0.34 & -7.67 $\pm$ 0.30 \\
SELFIES VAE BO & -8.34 $\pm$ 0.44 & -8.59 $\pm$ 0.40 & -7.15 $\pm$ 0.37 \\
SMILES GA & -8.25 $\pm$ 0.35 & -8.47 $\pm$ 0.45 & -7.04 $\pm$ 0.30 \\
SMILES LSTM HC & -8.87 $\pm$ 0.31 & -9.00 $\pm$ 0.44 & -7.46 $\pm$ 0.27 \\
SMILES-VAE-BO & -8.36 $\pm$ 0.43 & -8.60 $\pm$ 0.40 & -7.14 $\pm$ 0.37 \\
\bottomrule
\end{tabular}
\end{sc}
\end{small}
\end{center}
\end{table*}

\begin{table*}[h]
\caption{Top 100 Docking score for each target protein. } 
\label{top100-docking-table}
\vskip 0.15in
\begin{center}
\begin{small}
\begin{sc}
\begin{tabular}{cccccccccccccc}
\toprule
Model \ & 1iep & 3eml & 3ny8 & 4rlu \\
\midrule
3DSBDD & -8.01 $\pm$ 0.52 & -9.15 $\pm$ 0.42 & -8.60 $\pm$ 0.72 & -8.71 $\pm$ 0.53 \\
AutoGrow4 & \textbf{-11.80} $\pm$ 0.84 & \textbf{-12.14} $\pm$ 0.40 & \textbf{-10.84} $\pm$ 0.34 & \textbf{-10.73} $\pm$ 0.24 \\
Pocket2mol & -9.36 $\pm$ 0.36 & -11.30 $\pm$ 0.44 & -10.73 $\pm$ 0.60 & -9.95 $\pm$ 0.30 \\
PocketFlow & -11.20 $\pm$ 0.58 & -8.03 $\pm$ 0.60 & -7.69 $\pm$ 0.38 & -8.64 $\pm$ 0.43 \\
ResGen & -10.03 $\pm$ 0.44 & -6.43 $\pm$ 1.14 & -10.04 $\pm$ 0.42 & -10.27 $\pm$ 0.68 \\
DST & -9.72 $\pm$ 0.53 & -9.74 $\pm$ 0.40 & -9.77 $\pm$ 0.35 & -9.91 $\pm$ 0.44 \\
graph GA & -8.67 $\pm$ 0.65 & -8.80 $\pm$ 0.51 & -8.85 $\pm$ 0.55 & -9.04 $\pm$ 0.60 \\
MIMOSA & -9.74 $\pm$ 0.53 & -9.73 $\pm$ 0.40 & -9.77 $\pm$ 0.35 & -9.90 $\pm$ 0.43 \\
MolDQN & -6.03 $\pm$ 0.36 & -5.85 $\pm$ 0.36 & -5.94 $\pm$ 0.57 & -5.82 $\pm$ 0.44 \\
Pasithea & -9.72 $\pm$ 0.52 & -9.65 $\pm$ 0.32 & -9.77 $\pm$ 0.43 & -9.88 $\pm$ 0.48 \\
REINVENT & -7.59 $\pm$ 1.23 & -7.61 $\pm$ 1.50 & -7.81 $\pm$ 1.16 & -7.80 $\pm$ 1.16 \\
screening & -9.84 $\pm$ 0.50 & -9.76 $\pm$ 0.51 & -9.80 $\pm$ 0.44 & -9.91 $\pm$ 0.43 \\
SELFIES VAE BO & -8.64 $\pm$ 0.75 & -8.75 $\pm$ 0.59 & -8.86 $\pm$ 0.59 & -8.87 $\pm$ 0.55 \\
SMILES GA & -8.50 $\pm$ 0.50 & -8.23 $\pm$ 0.57 & -8.74 $\pm$ 0.54 & -8.77 $\pm$ 0.38 \\
SMILES LSTM HC & -9.14 $\pm$ 0.58 & -9.44 $\pm$ 0.44 & -9.20 $\pm$ 0.52 & -9.35 $\pm$ 0.53 \\
SMILES-VAE-BO & -8.64 $\pm$ 0.67 & -8.75 $\pm$ 0.61 & -8.86 $\pm$ 0.58 & -8.88 $\pm$ 0.56 \\
\bottomrule
\end{tabular}
\end{sc}
\end{small}
\end{center}

\begin{center}
\begin{small}
\begin{sc}
\begin{tabular}{cccccccccccccc}
\toprule
Model \ & 4unn & 5mo4 & 7l11 \\
\midrule
3DSBDD & -6.71 $\pm$ 0.86 & -7.50 $\pm$ 0.62 & -7.86 $\pm$ 0.29 \\
AutoGrow4 & -10.81 $\pm$ 0.13 & \textbf{-10.10} $\pm$ 0.16 & -8.32 $\pm$ 0.22 \\
Pocket2mol & \textbf{-11.22} $\pm$ 0.45 & -9.20 $\pm$ 0.48 & \textbf{-8.78} $\pm$ 0.44 \\
PocketFlow & -6.58 $\pm$ 0.56 & -6.70 $\pm$ 0.49 & -7.62 $\pm$ 0.33 \\
ResGen & -8.13 $\pm$ 0.64 & -8.88 $\pm$ 0.66 & -7.97 $\pm$ 0.39 \\
DST & -8.83 $\pm$ 0.39 & -9.03 $\pm$ 0.41 & -7.55 $\pm$ 0.36 \\
graph GA & -8.03 $\pm$ 0.56 & -8.22 $\pm$ 0.53 & -6.87 $\pm$ 0.40 \\
MIMOSA & -8.81 $\pm$ 0.39 & -9.05 $\pm$ 0.41 & -7.56 $\pm$ 0.37 \\
MolDQN & -5.22 $\pm$ 0.33 & -5.47 $\pm$ 0.40 & -5.51 $\pm$ 0.61 \\
Pasithea & -8.72 $\pm$ 0.43 & -8.94 $\pm$ 0.38 & -7.46 $\pm$ 0.29 \\
REINVENT & -7.14 $\pm$ 0.97 & -7.16 $\pm$ 1.06 & -6.06 $\pm$ 0.75 \\
screening & -8.85 $\pm$ 0.43 & -9.11 $\pm$ 0.36 & -7.44 $\pm$ 0.32 \\
SELFIES VAE BO & -7.93 $\pm$ 0.52 & -8.19 $\pm$ 0.50 & -6.81 $\pm$ 0.44 \\
SMILES GA & -7.97 $\pm$ 0.37 & -8.14 $\pm$ 0.47 & -6.83 $\pm$ 0.31 \\
SMILES LSTM HC & -8.53 $\pm$ 0.42 & -8.64 $\pm$ 0.49 & -7.20 $\pm$ 0.33 \\
SMILES-VAE-BO & -7.95 $\pm$ 0.52 & -8.20 $\pm$ 0.50 & -6.81 $\pm$ 0.43 \\

\bottomrule
\end{tabular}
\end{sc}
\end{small}
\end{center}
\end{table*}

\begin{table*}[h]
\caption{Top 1 LogP score for each target protein. } 
\label{top1-logp-table}
\vskip 0.15in
\begin{center}
\begin{small}
\begin{sc}
\begin{tabular}{cccccccccccccc}
\toprule
Model \ & 1iep & 3eml & 3ny8 & 4rlu \\
\midrule
3DSBDD & 1.23 & 2.94 & 2.66 & 0.60 \\
AutoGrow4 & 3.47 & 2.95 & 2.63 & 2.86 \\
Pocket2mol & 2.32 & 3.83 & 4.14 & 3.50 \\
PocketFlow & 6.10 & 4.86 & 2.57 & 4.38 \\
ResGen & 3.91 & 10.72 & 5.47 & 3.50 \\
DST & 3.72 & 3.72 & 3.72 & 3.72 \\
graph GA & 3.06 & 3.32 & 3.67 & 3.70 \\
MIMOSA & 3.72 & 3.72 & 3.72 & 3.72 \\
MolDQN & 0.36 & 0.55 & 0.36 & 0.33 \\
Pasithea & 3.22 & 3.22 & 3.22 & 3.22 \\
reinvent & 3.90 & 3.90 & 3.90 & 3.90 \\
screening & 3.67 & 3.67 & 3.67 & 3.67 \\
SELFIES VAE BO & 3.22 & 3.22 & 3.22 & 3.22 \\
smiles GA & 3.00 & 3.00 & 3.40 & 3.00 \\
smiles lstm hc & 6.51 & 6.51 & 6.51 & 6.51 \\
SMILES-VAE-BO & 3.22 & 3.22 & 3.22 & 3.22 \\
\bottomrule
\end{tabular}
\end{sc}
\end{small}
\end{center}

\begin{center}
\begin{small}
\begin{sc}
\begin{tabular}{cccccccccccccc}
\toprule
Model \ & 4unn & 5mo4 & 7l11 \\
\midrule
3DSBDD & 2.69 & 2.10 & 3.36 \\
AutoGrow4 & 2.72 & 2.93 & 2.81 \\
Pocket2mol & 2.86 & 3.66 & 3.92 \\
PocketFlow & 2.16 & 2.26 & 5.21 \\
ResGen & 3.40 & 3.54 & 3.23 \\
DST & 3.72 & 3.72 & 3.72 \\
graph GA & 3.13 & 3.01 & 3.76 \\
MIMOSA & 3.72 & 3.72 & 3.72 \\
molDQN & -0.26 & 1.00 & 0.50 \\
Pasithea & 3.22 & 3.22 & 3.22 \\
REINVENT & 3.90 & 3.90 & 3.90 \\
screening & 3.67 & 3.67 & 3.67 \\
selfies VAE BO & 3.22 & 3.22 & 3.22 \\
SMILES GA & 3.40 & 3.00 & 3.00 \\
smiles LSTM HC & 6.51 & 6.51 & 6.51 \\
SMILES-VAE-BO & 3.22 & 3.22 & 3.22 \\

\bottomrule
\end{tabular}
\end{sc}
\end{small}
\end{center}
\end{table*}

\begin{table*}[h]
\caption{Top 10 LogP score for each target protein. } 
\label{top10-logp-table}
\vskip 0.15in
\begin{center}
\begin{small}
\begin{sc}
\begin{tabular}{cccccccccccccc}
\toprule
Model \ & 1iep & 3eml & 3ny8 & 4rlu \\
\midrule
3DSBDD & 0.16 $\pm$ 0.37 & 1.59 $\pm$ 0.67 & 1.89 $\pm$ 0.51 & -0.18 $\pm$ 0.60 \\
AutoGrow4 & 3.36 $\pm$ 0.05 & 2.72 $\pm$ 0.13 & 2.63 $\pm$ 0.00 & 2.86 $\pm$ 0.00 \\
Pocket2mol & 2.10 $\pm$ 0.13 & 3.59 $\pm$ 0.15 & 3.66 $\pm$ 0.22 & 3.15 $\pm$ 0.19 \\
PocketFlow & 5.81 $\pm$ 0.17 & 2.21 $\pm$ 0.92 & 2.01 $\pm$ 0.32 & 3.79 $\pm$ 0.59 \\
ResGen & 3.58 $\pm$ 0.24 & 6.62 $\pm$ 1.55 & 4.91 $\pm$ 0.32 & 3.05 $\pm$ 0.25 \\
DST & 3.13 $\pm$ 0.21 & 3.13 $\pm$ 0.21 & 3.13 $\pm$ 0.21 & 3.13 $\pm$ 0.21 \\
graph GA & 2.79 $\pm$ 0.20 & 2.91 $\pm$ 0.23 & 3.14 $\pm$ 0.28 & 3.00 $\pm$ 0.32 \\
MIMOSA & 3.13 $\pm$ 0.21 & 3.13 $\pm$ 0.21 & 3.13 $\pm$ 0.21 & 3.13 $\pm$ 0.21 \\
molDQN & -0.75 $\pm$ 0.49 & -0.52 $\pm$ 0.48 & -0.75 $\pm$ 0.50 & -0.35 $\pm$ 0.29 \\
Pasithea & 3.03 $\pm$ 0.09 & 3.03 $\pm$ 0.09 & 3.03 $\pm$ 0.09 & 3.03 $\pm$ 0.09 \\
REINVENT & 2.50 $\pm$ 0.51 & 2.47 $\pm$ 0.53 & 2.50 $\pm$ 0.51 & 2.47 $\pm$ 0.53 \\
screening & 3.30 $\pm$ 0.18 & 3.30 $\pm$ 0.18 & 3.30 $\pm$ 0.18 & 3.30 $\pm$ 0.18 \\
SELFIES VAE BO & 2.68 $\pm$ 0.33 & 2.68 $\pm$ 0.33 & 2.68 $\pm$ 0.33 & 2.68 $\pm$ 0.33 \\
SMILES GA & 2.32 $\pm$ 0.30 & 2.02 $\pm$ 0.36 & 2.52 $\pm$ 0.43 & 2.17 $\pm$ 0.29 \\
SMILES LSTM HC & 4.72 $\pm$ 0.85 & 4.72 $\pm$ 0.85 & 4.72 $\pm$ 0.85 & 4.72 $\pm$ 0.85 \\
SMILES-VAE-BO & 2.68 $\pm$ 0.33 & 2.68 $\pm$ 0.33 & 2.68 $\pm$ 0.33 & 2.68 $\pm$ 0.33 \\

\bottomrule
\end{tabular}
\end{sc}
\end{small}
\end{center}

\begin{center}
\begin{small}
\begin{sc}
\begin{tabular}{cccccccccccccc}
\toprule
Model \ & 4unn & 5mo4 & 7l11 \\
\midrule
3DSBDD & 2.02 $\pm$ 0.28 & 1.40 $\pm$ 0.39 & 2.46 $\pm$ 0.59 \\
AutoGrow4 & 2.68 $\pm$ 0.05 & 2.86 $\pm$ 0.06 & 2.77 $\pm$ 0.02 \\
Pocket2mol & 2.56 $\pm$ 0.16 & 3.23 $\pm$ 0.22 & 3.46 $\pm$ 0.19 \\
PocketFlow & 1.44 $\pm$ 0.28 & 1.00 $\pm$ 0.47 & 4.79 $\pm$ 0.26 \\
ResGen & 2.19 $\pm$ 0.60 & 2.71 $\pm$ 0.49 & 2.74 $\pm$ 0.27 \\
DST & 3.13 $\pm$ 0.21 & 3.13 $\pm$ 0.21 & 3.13 $\pm$ 0.21 \\
graph GA & 2.84 $\pm$ 0.19 & 2.71 $\pm$ 0.20 & 2.99 $\pm$ 0.29 \\
MIMOSA & 3.13 $\pm$ 0.21 & 3.13 $\pm$ 0.21 & 3.13 $\pm$ 0.21 \\
MolDQN & -0.95 $\pm$ 0.43 & -0.12 $\pm$ 0.54 & -0.23 $\pm$ 0.48 \\
Pasithea & 3.03 $\pm$ 0.09 & 3.03 $\pm$ 0.09 & 3.03 $\pm$ 0.09 \\
REINVENT & 2.50 $\pm$ 0.51 & 2.50 $\pm$ 0.51 & 2.50 $\pm$ 0.51 \\
screening & 3.30 $\pm$ 0.18 & 3.30 $\pm$ 0.18 & 3.30 $\pm$ 0.18 \\
SELFIES VAE BO & 2.68 $\pm$ 0.33 & 2.68 $\pm$ 0.33 & 2.68 $\pm$ 0.33 \\
SMILES GA & 2.48 $\pm$ 0.44 & 2.34 $\pm$ 0.30 & 2.14 $\pm$ 0.34 \\
SMILES LSTM HC & 4.72 $\pm$ 0.85 & 4.72 $\pm$ 0.85 & 4.72 $\pm$ 0.85 \\
SMILES VAE BO & 2.68 $\pm$ 0.33 & 2.68 $\pm$ 0.33 & 2.68 $\pm$ 0.33 \\
\bottomrule
\end{tabular}
\end{sc}
\end{small}
\end{center}
\end{table*}

\begin{table*}[h]
\caption{Top 50 LogP score for each target protein. } 
\label{top50-logp-table}
\vskip 0.15in
\begin{center}
\begin{small}
\begin{sc}
\begin{tabular}{cccccccccccccc}
\toprule
Model \ & 1iep & 3eml & 3ny8 & 4rlu \\
\midrule
3DSBDD & 0.03 $\pm$ 0.18 & 0.56 $\pm$ 0.64 & 0.53 $\pm$ 0.78 & -2.01 $\pm$ 1.00 \\
AutoGrow4 & 2.96 $\pm$ 0.22 & 2.56 $\pm$ 0.15 & 2.49 $\pm$ 0.13 & 2.67 $\pm$ 0.16 \\
Pocket2mol & 1.80 $\pm$ 0.18 & 3.12 $\pm$ 0.31 & 3.20 $\pm$ 0.30 & 2.76 $\pm$ 0.25 \\
PocketFlow & 5.38 $\pm$ 0.29 & 1.20 $\pm$ 0.70 & 0.98 $\pm$ 0.66 & 2.19 $\pm$ 0.92 \\
ResGen & 2.99 $\pm$ 0.37 & 5.22 $\pm$ 1.02 & 3.99 $\pm$ 0.54 & 2.20 $\pm$ 0.51 \\
DST & 2.78 $\pm$ 0.25 & 2.78 $\pm$ 0.25 & 2.78 $\pm$ 0.25 & 2.78 $\pm$ 0.25 \\
graph GA & 2.06 $\pm$ 0.42 & 2.18 $\pm$ 0.44 & 2.22 $\pm$ 0.55 & 2.18 $\pm$ 0.49 \\
mimosa & 2.78 $\pm$ 0.25 & 2.78 $\pm$ 0.25 & 2.78 $\pm$ 0.25 & 2.78 $\pm$ 0.25 \\
moldqn & -1.73 $\pm$ 0.60 & -1.51 $\pm$ 0.59 & -1.68 $\pm$ 0.58 & -1.20 $\pm$ 0.53 \\
Pasithea & 2.69 $\pm$ 0.25 & 2.69 $\pm$ 0.25 & 2.69 $\pm$ 0.25 & 2.69 $\pm$ 0.25 \\
reinvent & 1.50 $\pm$ 0.65 & 1.42 $\pm$ 0.69 & 1.46 $\pm$ 0.69 & 1.46 $\pm$ 0.66 \\
screening & 2.85 $\pm$ 0.27 & 2.85 $\pm$ 0.27 & 2.85 $\pm$ 0.27 & 2.85 $\pm$ 0.27 \\
selfies vae bo & 1.99 $\pm$ 0.42 & 1.99 $\pm$ 0.42 & 1.99 $\pm$ 0.42 & 1.99 $\pm$ 0.42 \\
smiles ga & 1.72 $\pm$ 0.39 & 1.25 $\pm$ 0.48 & 1.81 $\pm$ 0.47 & 1.73 $\pm$ 0.32 \\
smiles lstm hc & 3.51 $\pm$ 0.77 & 3.51 $\pm$ 0.77 & 3.51 $\pm$ 0.77 & 3.51 $\pm$ 0.77 \\
smiles vae bo & 1.99 $\pm$ 0.42 & 1.99 $\pm$ 0.42 & 1.99 $\pm$ 0.42 & 1.99 $\pm$ 0.42 \\

\bottomrule
\end{tabular}
\end{sc}
\end{small}
\end{center}

\begin{center}
\begin{small}
\begin{sc}
\begin{tabular}{cccccccccccccc}
\toprule
Model \ & 4unn & 5mo4 & 7l11 \\
\midrule
3DSBDD & 0.90 $\pm$ 0.77 & 0.30 $\pm$ 0.58 & 0.73 $\pm$ 0.98 \\
AutoGrow4 & 2.63 $\pm$ 0.04 & 2.70 $\pm$ 0.14 & 2.40 $\pm$ 0.31 \\
Pocket2mol & 2.12 $\pm$ 0.30 & 2.80 $\pm$ 0.28 & 2.96 $\pm$ 0.32 \\
PocketFlow & 0.61 $\pm$ 0.53 & 0.57 $\pm$ 0.31 & 4.06 $\pm$ 0.44 \\
ResGen & 0.81 $\pm$ 0.91 & 1.68 $\pm$ 0.62 & 1.97 $\pm$ 0.47 \\
DST & 2.78 $\pm$ 0.25 & 2.78 $\pm$ 0.25 & 2.78 $\pm$ 0.25 \\
graph ga & 2.15 $\pm$ 0.44 & 2.09 $\pm$ 0.37 & 2.27 $\pm$ 0.46 \\
mimosa & 2.78 $\pm$ 0.25 & 2.78 $\pm$ 0.25 & 2.78 $\pm$ 0.25 \\
moldqn & -1.88 $\pm$ 0.57 & -1.46 $\pm$ 0.79 & -1.34 $\pm$ 0.66 \\
Pasithea & 2.69 $\pm$ 0.25 & 2.69 $\pm$ 0.25 & 2.69 $\pm$ 0.25 \\
reinvent & 1.47 $\pm$ 0.68 & 1.48 $\pm$ 0.67 & 1.49 $\pm$ 0.66 \\
screening & 2.85 $\pm$ 0.27 & 2.85 $\pm$ 0.27 & 2.85 $\pm$ 0.27 \\
selfies vae bo & 1.99 $\pm$ 0.42 & 1.99 $\pm$ 0.42 & 1.99 $\pm$ 0.42 \\
smiles ga & 1.77 $\pm$ 0.46 & 1.70 $\pm$ 0.40 & 1.50 $\pm$ 0.42 \\
smiles lstm hc & 3.51 $\pm$ 0.77 & 3.51 $\pm$ 0.77 & 3.51 $\pm$ 0.77 \\
smiles vae bo & 1.99 $\pm$ 0.42 & 1.99 $\pm$ 0.42 & 1.99 $\pm$ 0.42 \\
\bottomrule
\end{tabular}
\end{sc}
\end{small}
\end{center}
\end{table*}

\begin{table*}[h]
\caption{Top 100 LogP score for each target protein. } 
\label{top100-logp-table}
\vskip 0.15in
\begin{center}
\begin{small}
\begin{sc}
\begin{tabular}{cccccccccccccc}
\toprule
Model \ & 1iep & 3eml & 3ny8 & 4rlu \\
\midrule
3DSBDD & 0.02 $\pm$ 0.13 & 0.28 $\pm$ 0.53 & 0.26 $\pm$ 0.61 & -2.58 $\pm$ 0.92 \\
AutoGrow4 & 2.83 $\pm$ 0.21 & 2.37 $\pm$ 0.23 & 2.20 $\pm$ 0.33 & 2.31 $\pm$ 0.40 \\
Pocket2mol & 1.61 $\pm$ 0.24 & 2.79 $\pm$ 0.40 & 2.91 $\pm$ 0.37 & 2.50 $\pm$ 0.32 \\
PocketFlow & 5.02 $\pm$ 0.43 & 0.57 $\pm$ 0.82 & 0.40 $\pm$ 0.76 & 1.42 $\pm$ 1.02 \\
ResGen & 2.65 $\pm$ 0.44 & 4.52 $\pm$ 1.05 & 3.59 $\pm$ 0.56 & 1.72 $\pm$ 0.61 \\
DST & 2.51 $\pm$ 0.32 & 2.51 $\pm$ 0.32 & 2.51 $\pm$ 0.32 & 2.51 $\pm$ 0.32 \\
graph ga & 1.69 $\pm$ 0.49 & 1.78 $\pm$ 0.52 & 1.76 $\pm$ 0.62 & 1.76 $\pm$ 0.56 \\
mimosa & 2.51 $\pm$ 0.32 & 2.51 $\pm$ 0.32 & 2.51 $\pm$ 0.32 & 2.51 $\pm$ 0.32 \\
moldqn & -2.21 $\pm$ 0.66 & -1.97 $\pm$ 0.63 & -2.19 $\pm$ 0.67 & -1.72 $\pm$ 0.66 \\
Pasithea & 2.42 $\pm$ 0.33 & 2.42 $\pm$ 0.33 & 2.42 $\pm$ 0.33 & 2.42 $\pm$ 0.33 \\
reinvent & 0.15 $\pm$ 1.82 & 0.02 $\pm$ 1.89 & 0.05 $\pm$ 1.90 & 0.07 $\pm$ 1.88 \\
screening & 2.60 $\pm$ 0.32 & 2.60 $\pm$ 0.32 & 2.60 $\pm$ 0.32 & 2.60 $\pm$ 0.32 \\
selfies vae bo & 1.43 $\pm$ 0.68 & 1.43 $\pm$ 0.68 & 1.43 $\pm$ 0.68 & 1.43 $\pm$ 0.68 \\
smiles ga & 1.35 $\pm$ 0.47 & 0.80 $\pm$ 0.57 & 1.41 $\pm$ 0.52 & 1.35 $\pm$ 0.45 \\
smiles lstm hc & 2.96 $\pm$ 0.78 & 2.96 $\pm$ 0.78 & 2.96 $\pm$ 0.78 & 2.96 $\pm$ 0.78 \\
smiles vae bo & 1.43 $\pm$ 0.68 & 1.43 $\pm$ 0.68 & 1.43 $\pm$ 0.68 & 1.43 $\pm$ 0.68 \\
\bottomrule
\end{tabular}
\end{sc}
\end{small}
\end{center}

\begin{center}
\begin{small}
\begin{sc}
\begin{tabular}{cccccccccccccc}
\toprule
Model \ & 4unn & 5mo4 & 7l11 \\
\midrule
3DSBDD & 0.45 $\pm$ 0.70 & 0.15 $\pm$ 0.44 & -0.05 $\pm$ 1.22 \\
AutoGrow4 & 2.46 $\pm$ 0.21 & 2.43 $\pm$ 0.30 & 2.10 $\pm$ 0.38 \\
Pocket2mol & 1.80 $\pm$ 0.39 & 2.53 $\pm$ 0.35 & 2.50 $\pm$ 0.54 \\
PocketFlow & 0.04 $\pm$ 0.69 & 0.19 $\pm$ 0.46 & 3.56 $\pm$ 0.62 \\
ResGen & 0.00 $\pm$ 1.05 & 1.20 $\pm$ 0.66 & 1.48 $\pm$ 0.61 \\
DST & 2.51 $\pm$ 0.32 & 2.51 $\pm$ 0.32 & 2.51 $\pm$ 0.32 \\
graph ga & 1.70 $\pm$ 0.56 & 1.70 $\pm$ 0.48 & 1.84 $\pm$ 0.55 \\
mimosa & 2.51 $\pm$ 0.32 & 2.51 $\pm$ 0.32 & 2.51 $\pm$ 0.32 \\
moldqn & -2.34 $\pm$ 0.62 & -2.07 $\pm$ 0.84 & -1.93 $\pm$ 0.77 \\
Pasithea & 2.42 $\pm$ 0.33 & 2.42 $\pm$ 0.33 & 2.42 $\pm$ 0.33 \\
reinvent & 0.05 $\pm$ 1.90 & 0.08 $\pm$ 1.89 & 0.09 $\pm$ 1.89 \\
screening & 2.60 $\pm$ 0.32 & 2.60 $\pm$ 0.32 & 2.60 $\pm$ 0.32 \\
selfies vae bo & 1.43 $\pm$ 0.68 & 1.43 $\pm$ 0.68 & 1.43 $\pm$ 0.68 \\
smiles ga & 1.39 $\pm$ 0.51 & 1.37 $\pm$ 0.44 & 1.10 $\pm$ 0.52 \\
smiles lstm hc & 2.96 $\pm$ 0.78 & 2.96 $\pm$ 0.78 & 2.96 $\pm$ 0.78 \\
smiles vae bo & 1.43 $\pm$ 0.68 & 1.43 $\pm$ 0.68 & 1.43 $\pm$ 0.68 \\
\bottomrule
\end{tabular}
\end{sc}
\end{small}
\end{center}
\end{table*}

\begin{table*}[h]
\caption{Top 1 QED score for each target protein. } 
\label{top1-qed-table}
\vskip 0.15in
\begin{center}
\begin{small}
\begin{sc}
\begin{tabular}{cccccccccccccc}
\toprule
Model \ & 1iep & 3eml & 3ny8 & 4rlu \\
\midrule
3DSBDD & 0.81 & 0.93 & \textbf{0.95} & 0.85 \\
AutoGrow4 & 0.86 & 0.85 & 0.92 & 0.84 \\
Pocket2mol & 0.92 & 0.91 & 0.93 & 0.92 \\
PocketFlow & 0.92 & 0.86 & 0.81 & 0.88 \\
ResGen & 0.92 & 0.80 & 0.93 & 0.94 \\
dst & \textbf{0.95} & \textbf{0.95} & 0.95 & \textbf{0.95} \\
graph ga & 0.94 & 0.94 & 0.94 & 0.94 \\
mimosa & \textbf{0.95} & \textbf{0.95} & 0.95 & \textbf{0.95} \\
moldqn & 0.52 & 0.65 & 0.62 & 0.67 \\
Pasithea & \textbf{0.95} & \textbf{0.95} & 0.95 & \textbf{0.95} \\
reinvent & 0.95 & 0.95 & 0.95 & 0.95 \\
screening & 0.95 & 0.95 & 0.95 & 0.95 \\
selfies vae bo & 0.94 & 0.94 & 0.94 & 0.94 \\
smiles ga & 0.93 & 0.93 & 0.94 & 0.94 \\
smiles lstm hc & 0.94 & 0.94 & 0.94 & 0.94 \\
smiles vae bo & 0.94 & 0.94 & 0.94 & 0.94 \\
\bottomrule
\end{tabular}
\end{sc}
\end{small}
\end{center}

\begin{center}
\begin{small}
\begin{sc}
\begin{tabular}{cccccccccccccc}
\toprule
Model \ & 4unn & 5mo4 & 7l11 \\
\midrule
3DSBDD & 0.88 & 0.80 & 0.90 \\
AutoGrow4 & 0.89 & 0.87 & 0.84 \\
Pocket2mol & 0.88 & 0.90 & 0.94 \\
PocketFlow & 0.83 & 0.80 & 0.91 \\
ResGen & 0.93 & 0.88 & 0.95 \\
DST & \textbf{0.95} & \textbf{0.95} & 0.95 \\
graph ga & 0.94 & 0.94 & \textbf{0.95} \\
mimosa & \textbf{0.95} & \textbf{0.95} & 0.95 \\
moldqn & 0.67 & 0.52 & 0.70 \\
Pasithea & \textbf{0.95} & \textbf{0.95} & 0.95 \\
reinvent & 0.95 & 0.95 & 0.95 \\
screening & 0.95 & 0.95 & 0.95 \\
selfies vae bo & 0.94 & 0.94 & 0.94 \\
smiles ga & 0.93 & 0.93 & 0.94 \\
smiles lstm hc & 0.94 & 0.94 & 0.94 \\
smiles vae bo & 0.94 & 0.94 & 0.94 \\
\bottomrule
\end{tabular}
\end{sc}
\end{small}
\end{center}
\end{table*}

\begin{table*}[h]
\caption{Top 10 QED score for each target protein. } 
\label{top10-qed-table}
\vskip 0.15in
\begin{center}
\begin{small}
\begin{sc}
\begin{tabular}{cccccccccccccc}
\toprule
Model \ & 1iep & 3eml & 3ny8 & 4rlu \\
\midrule
3DSBDD & 0.77 $\pm$ 0.02 & 0.83 $\pm$ 0.05 & 0.88 $\pm$ 0.03 & 0.81 $\pm$ 0.02 \\
AutoGrow4 & 0.83 $\pm$ 0.01 & 0.79 $\pm$ 0.04 & 0.92 $\pm$ 0.00 & 0.82 $\pm$ 0.01 \\
Pocket2mol & 0.90 $\pm$ 0.01 & 0.89 $\pm$ 0.01 & 0.92 $\pm$ 0.01 & 0.90 $\pm$ 0.01 \\
PocketFlow & 0.89 $\pm$ 0.01 & 0.81 $\pm$ 0.03 & 0.77 $\pm$ 0.03 & 0.79 $\pm$ 0.03 \\
ResGen & 0.91 $\pm$ 0.01 & 0.73 $\pm$ 0.03 & 0.90 $\pm$ 0.02 & 0.93 $\pm$ 0.00 \\
DST & 0.94 $\pm$ 0.00 & 0.94 $\pm$ 0.00 & 0.94 $\pm$ 0.00 & 0.94 $\pm$ 0.00 \\
graph ga & 0.93 $\pm$ 0.01 & 0.93 $\pm$ 0.01 & 0.93 $\pm$ 0.01 & 0.92 $\pm$ 0.01 \\
mimosa & 0.94 $\pm$ 0.00 & 0.94 $\pm$ 0.00 & 0.94 $\pm$ 0.00 & 0.94 $\pm$ 0.00 \\
moldqn & 0.49 $\pm$ 0.02 & 0.55 $\pm$ 0.04 & 0.51 $\pm$ 0.05 & 0.60 $\pm$ 0.05 \\
Pasithea & 0.94 $\pm$ 0.00 & 0.94 $\pm$ 0.00 & 0.94 $\pm$ 0.00 & 0.94 $\pm$ 0.00 \\
reinvent & 0.91 $\pm$ 0.02 & 0.91 $\pm$ 0.02 & 0.91 $\pm$ 0.02 & 0.91 $\pm$ 0.02 \\
screening & \textbf{0.94} $\pm$ 0.00 & \textbf{0.94} $\pm$ 0.00 & \textbf{0.94} $\pm$ 0.00 & \textbf{0.94} $\pm$ 0.00 \\
selfies vae bo & 0.91 $\pm$ 0.02 & 0.91 $\pm$ 0.02 & 0.91 $\pm$ 0.02 & 0.91 $\pm$ 0.02 \\
smiles ga & 0.92 $\pm$ 0.01 & 0.93 $\pm$ 0.01 & 0.93 $\pm$ 0.00 & 0.93 $\pm$ 0.01 \\
smiles lstm hc & 0.92 $\pm$ 0.01 & 0.92 $\pm$ 0.01 & 0.92 $\pm$ 0.01 & 0.92 $\pm$ 0.01 \\
smiles vae bo & 0.91 $\pm$ 0.02 & 0.91 $\pm$ 0.02 & 0.91 $\pm$ 0.02 & 0.91 $\pm$ 0.02 \\
\bottomrule
\end{tabular}
\end{sc}
\end{small}
\end{center}

\begin{center}
\begin{small}
\begin{sc}
\begin{tabular}{cccccccccccccc}
\toprule
Model \ & 4unn & 5mo4 & 7l11 \\
\midrule
3DSBDD & 0.82 $\pm$ 0.03 & 0.76 $\pm$ 0.02 & 0.86 $\pm$ 0.02 \\
AutoGrow4 & 0.87 $\pm$ 0.01 & 0.83 $\pm$ 0.03 & 0.80 $\pm$ 0.02 \\
Pocket2mol & 0.84 $\pm$ 0.02 & 0.88 $\pm$ 0.01 & 0.94 $\pm$ 0.01 \\
PocketFlow & 0.74 $\pm$ 0.04 & 0.74 $\pm$ 0.03 & 0.90 $\pm$ 0.01 \\
ResGen & 0.85 $\pm$ 0.04 & 0.84 $\pm$ 0.01 & 0.92 $\pm$ 0.02 \\
DST & 0.94 $\pm$ 0.00 & 0.94 $\pm$ 0.00 & 0.94 $\pm$ 0.00 \\
graph ga & 0.93 $\pm$ 0.01 & 0.93 $\pm$ 0.01 & 0.93 $\pm$ 0.01 \\
mimosa & 0.94 $\pm$ 0.00 & 0.94 $\pm$ 0.00 & 0.94 $\pm$ 0.00 \\
moldqn & 0.57 $\pm$ 0.04 & 0.48 $\pm$ 0.02 & 0.54 $\pm$ 0.07 \\
Pasithea & 0.94 $\pm$ 0.00 & 0.94 $\pm$ 0.00 & 0.94 $\pm$ 0.00 \\
reinvent & 0.91 $\pm$ 0.02 & 0.91 $\pm$ 0.02 & 0.91 $\pm$ 0.02 \\
screening & \textbf{0.94} $\pm$ 0.00 & \textbf{0.94} $\pm$ 0.00 & \textbf{0.94} $\pm$ 0.00 \\
selfies vae bo & 0.91 $\pm$ 0.02 & 0.91 $\pm$ 0.02 & 0.91 $\pm$ 0.02 \\
smiles ga & 0.93 $\pm$ 0.00 & 0.93 $\pm$ 0.00 & 0.92 $\pm$ 0.01 \\
smiles lstm hc & 0.92 $\pm$ 0.01 & 0.92 $\pm$ 0.01 & 0.92 $\pm$ 0.01 \\
smiles vae bo & 0.91 $\pm$ 0.02 & 0.91 $\pm$ 0.02 & 0.91 $\pm$ 0.02 \\
\bottomrule
\end{tabular}
\end{sc}
\end{small}
\end{center}
\end{table*}

\begin{table*}[h]
\caption{Top 50 QED score for each target protein. } 
\label{top50-qed-table}
\vskip 0.15in
\begin{center}
\begin{small}
\begin{sc}
\begin{tabular}{cccccccccccccc}
\toprule
Model \ & 1iep & 3eml & 3ny8 & 4rlu \\
\midrule
3DSBDD & 0.69 $\pm$ 0.05 & 0.71 $\pm$ 0.07 & 0.79 $\pm$ 0.05 & 0.77 $\pm$ 0.02 \\
AutoGrow4 & 0.78 $\pm$ 0.03 & 0.73 $\pm$ 0.04 & 0.88 $\pm$ 0.03 & 0.76 $\pm$ 0.04 \\
Pocket2mol & 0.87 $\pm$ 0.02 & 0.85 $\pm$ 0.03 & 0.88 $\pm$ 0.03 & 0.87 $\pm$ 0.02 \\
PocketFlow & 0.85 $\pm$ 0.03 & 0.71 $\pm$ 0.06 & 0.69 $\pm$ 0.04 & 0.69 $\pm$ 0.06 \\
ResGen & 0.86 $\pm$ 0.04 & 0.65 $\pm$ 0.05 & 0.85 $\pm$ 0.03 & 0.90 $\pm$ 0.02 \\
DST & 0.92 $\pm$ 0.01 & 0.92 $\pm$ 0.01 & 0.92 $\pm$ 0.01 & 0.92 $\pm$ 0.01 \\
graph ga & 0.89 $\pm$ 0.02 & 0.89 $\pm$ 0.03 & 0.89 $\pm$ 0.03 & 0.88 $\pm$ 0.03 \\
mimosa & 0.92 $\pm$ 0.01 & 0.92 $\pm$ 0.01 & 0.92 $\pm$ 0.01 & 0.92 $\pm$ 0.01 \\
moldqn & 0.43 $\pm$ 0.03 & 0.48 $\pm$ 0.04 & 0.44 $\pm$ 0.04 & 0.49 $\pm$ 0.06 \\
Pasithea & 0.92 $\pm$ 0.01 & 0.92 $\pm$ 0.01 & 0.92 $\pm$ 0.01 & 0.92 $\pm$ 0.01 \\
reinvent & 0.84 $\pm$ 0.05 & 0.84 $\pm$ 0.05 & 0.84 $\pm$ 0.05 & 0.84 $\pm$ 0.05 \\
screening & \textbf{0.92} $\pm$ 0.01 & \textbf{0.92} $\pm$ 0.01 & \textbf{0.92} $\pm$ 0.01 & \textbf{0.92} $\pm$ 0.01 \\
selfies vae bo & 0.87 $\pm$ 0.03 & 0.87 $\pm$ 0.03 & 0.87 $\pm$ 0.03 & 0.87 $\pm$ 0.03 \\
smiles ga & 0.87 $\pm$ 0.03 & 0.87 $\pm$ 0.03 & 0.89 $\pm$ 0.02 & 0.89 $\pm$ 0.02 \\
smiles lstm hc & 0.88 $\pm$ 0.03 & 0.88 $\pm$ 0.03 & 0.88 $\pm$ 0.03 & 0.88 $\pm$ 0.03 \\
smiles vae bo & 0.87 $\pm$ 0.03 & 0.87 $\pm$ 0.03 & 0.87 $\pm$ 0.03 & 0.87 $\pm$ 0.03 \\
\bottomrule
\end{tabular}
\end{sc}
\end{small}
\end{center}

\begin{center}
\begin{small}
\begin{sc}
\begin{tabular}{cccccccccccccc}
\toprule
Model \ & 4unn & 5mo4 & 7l11 \\
\midrule
3DSBDD & 0.74 $\pm$ 0.06 & 0.64 $\pm$ 0.08 & 0.74 $\pm$ 0.07 \\
AutoGrow4 & 0.86 $\pm$ 0.01 & 0.75 $\pm$ 0.05 & 0.74 $\pm$ 0.04 \\
Pocket2mol & 0.80 $\pm$ 0.03 & 0.84 $\pm$ 0.03 & 0.91 $\pm$ 0.02 \\
PocketFlow & 0.67 $\pm$ 0.04 & 0.68 $\pm$ 0.04 & 0.85 $\pm$ 0.03 \\
ResGen & 0.76 $\pm$ 0.06 & 0.78 $\pm$ 0.04 & 0.85 $\pm$ 0.04 \\
DST & 0.92 $\pm$ 0.01 & 0.92 $\pm$ 0.01 & 0.92 $\pm$ 0.01 \\
graph ga & 0.89 $\pm$ 0.03 & 0.89 $\pm$ 0.02 & 0.89 $\pm$ 0.03 \\
mimosa & 0.92 $\pm$ 0.01 & 0.92 $\pm$ 0.01 & 0.92 $\pm$ 0.01 \\
moldqn & 0.49 $\pm$ 0.05 & 0.43 $\pm$ 0.03 & 0.45 $\pm$ 0.06 \\
Pasithea & 0.92 $\pm$ 0.01 & 0.92 $\pm$ 0.01 & 0.92 $\pm$ 0.01 \\
reinvent & 0.83 $\pm$ 0.06 & 0.83 $\pm$ 0.05 & 0.83 $\pm$ 0.05 \\
screening & \textbf{0.92} $\pm$ 0.01 & \textbf{0.92} $\pm$ 0.01 & \textbf{0.92} $\pm$ 0.01 \\
selfies vae bo & 0.87 $\pm$ 0.03 & 0.87 $\pm$ 0.03 & 0.87 $\pm$ 0.03 \\
smiles ga & 0.90 $\pm$ 0.02 & 0.89 $\pm$ 0.02 & 0.88 $\pm$ 0.03 \\
smiles lstm hc & 0.88 $\pm$ 0.03 & 0.88 $\pm$ 0.03 & 0.88 $\pm$ 0.03 \\
smiles vae bo & 0.87 $\pm$ 0.03 & 0.87 $\pm$ 0.03 & 0.87 $\pm$ 0.03 \\
\bottomrule
\end{tabular}
\end{sc}
\end{small}
\end{center}
\end{table*}

\begin{table*}[h]
\caption{Top 100 QED score for each target protein. } 
\label{top100-qed-table}
\vskip 0.15in
\begin{center}
\begin{small}
\begin{sc}
\begin{tabular}{cccccccccccccc}
\toprule
Model \ & 1iep & 3eml & 3ny8 & 4rlu \\
\midrule
3DSBDD & 0.64 $\pm$ 0.07 & 0.65 $\pm$ 0.08 & 0.73 $\pm$ 0.08 & 0.75 $\pm$ 0.03 \\
AutoGrow4 & 0.76 $\pm$ 0.03 & 0.67 $\pm$ 0.07 & 0.84 $\pm$ 0.05 & 0.72 $\pm$ 0.05 \\
Pocket2mol & 0.85 $\pm$ 0.03 & 0.82 $\pm$ 0.03 & 0.85 $\pm$ 0.04 & 0.86 $\pm$ 0.02 \\
PocketFlow & 0.80 $\pm$ 0.05 & 0.65 $\pm$ 0.07 & 0.65 $\pm$ 0.05 & 0.64 $\pm$ 0.07 \\
ResGen & 0.82 $\pm$ 0.05 & 0.60 $\pm$ 0.06 & 0.83 $\pm$ 0.04 & 0.86 $\pm$ 0.04 \\
DST & 0.91 $\pm$ 0.02 & 0.91 $\pm$ 0.02 & 0.91 $\pm$ 0.02 & 0.91 $\pm$ 0.02 \\
graph ga & 0.85 $\pm$ 0.04 & 0.85 $\pm$ 0.04 & 0.85 $\pm$ 0.04 & 0.85 $\pm$ 0.04 \\
mimosa & 0.91 $\pm$ 0.02 & 0.91 $\pm$ 0.02 & 0.91 $\pm$ 0.02 & 0.91 $\pm$ 0.02 \\
moldqn & 0.40 $\pm$ 0.04 & 0.44 $\pm$ 0.05 & 0.41 $\pm$ 0.05 & 0.45 $\pm$ 0.06 \\
Pasithea & 0.90 $\pm$ 0.02 & 0.90 $\pm$ 0.02 & 0.90 $\pm$ 0.02 & 0.90 $\pm$ 0.02 \\
reinvent & 0.73 $\pm$ 0.14 & 0.73 $\pm$ 0.14 & 0.73 $\pm$ 0.14 & 0.73 $\pm$ 0.14 \\
screening & \textbf{0.91} $\pm$ 0.02 & \textbf{0.91} $\pm$ 0.02 & \textbf{0.91} $\pm$ 0.02 & \textbf{0.91} $\pm$ 0.02 \\
selfies vae bo & 0.83 $\pm$ 0.05 & 0.83 $\pm$ 0.05 & 0.83 $\pm$ 0.05 & 0.83 $\pm$ 0.05 \\
smiles ga & 0.84 $\pm$ 0.04 & 0.83 $\pm$ 0.05 & 0.86 $\pm$ 0.04 & 0.86 $\pm$ 0.04 \\
smiles lstm hc & 0.85 $\pm$ 0.04 & 0.85 $\pm$ 0.04 & 0.85 $\pm$ 0.04 & 0.85 $\pm$ 0.04 \\
smiles vae bo & 0.83 $\pm$ 0.05 & 0.83 $\pm$ 0.05 & 0.83 $\pm$ 0.05 & 0.83 $\pm$ 0.05 \\
\bottomrule
\end{tabular}
\end{sc}
\end{small}
\end{center}

\begin{center}
\begin{small}
\begin{sc}
\begin{tabular}{cccccccccccccc}
\toprule
Model \ & 4unn & 5mo4 & 7l11 \\
\midrule
3DSBDD & 0.65 $\pm$ 0.10 & 0.54 $\pm$ 0.12 & 0.66 $\pm$ 0.10 \\
AutoGrow4 & 0.84 $\pm$ 0.02 & 0.71 $\pm$ 0.06 & 0.69 $\pm$ 0.06 \\
Pocket2mol & 0.77 $\pm$ 0.04 & 0.81 $\pm$ 0.03 & 0.88 $\pm$ 0.03 \\
PocketFlow & 0.63 $\pm$ 0.05 & 0.64 $\pm$ 0.05 & 0.82 $\pm$ 0.04 \\
ResGen & 0.70 $\pm$ 0.07 & 0.75 $\pm$ 0.05 & 0.81 $\pm$ 0.05 \\
DST & 0.91 $\pm$ 0.02 & 0.91 $\pm$ 0.02 & 0.91 $\pm$ 0.02 \\
graph ga & 0.85 $\pm$ 0.05 & 0.86 $\pm$ 0.04 & 0.85 $\pm$ 0.04 \\
mimosa & 0.91 $\pm$ 0.02 & 0.91 $\pm$ 0.02 & 0.91 $\pm$ 0.02 \\
moldqn & 0.44 $\pm$ 0.06 & 0.39 $\pm$ 0.04 & 0.41 $\pm$ 0.06 \\
Pasithea & 0.90 $\pm$ 0.02 & 0.90 $\pm$ 0.02 & 0.90 $\pm$ 0.02 \\
reinvent & 0.73 $\pm$ 0.14 & 0.73 $\pm$ 0.13 & 0.73 $\pm$ 0.14 \\
screening & \textbf{0.91} $\pm$ 0.02 & \textbf{0.91} $\pm$ 0.02 & \textbf{0.91} $\pm$ 0.02 \\
selfies vae bo & 0.83 $\pm$ 0.05 & 0.83 $\pm$ 0.05 & 0.83 $\pm$ 0.05 \\
smiles ga & 0.88 $\pm$ 0.03 & 0.86 $\pm$ 0.03 & 0.84 $\pm$ 0.04 \\
smiles lstm hc & 0.85 $\pm$ 0.04 & 0.85 $\pm$ 0.04 & 0.85 $\pm$ 0.04 \\
smiles vae bo & 0.83 $\pm$ 0.05 & 0.83 $\pm$ 0.05 & 0.83 $\pm$ 0.05 \\
\bottomrule
\end{tabular}
\end{sc}
\end{small}
\end{center}
\end{table*}

\begin{table*}[h]
\caption{Top 1 SA score for each target protein. } 
\label{top1-sa-table}
\vskip 0.15in
\begin{center}
\begin{small}
\begin{sc}
\begin{tabular}{cccccccccccccc}
\toprule
Model \ & 1iep & 3eml & 3ny8 & 4rlu \\
\midrule
3DSBDD & 1.37 & \textbf{1.00} & \textbf{1.00} & 1.99 \\
AutoGrow4 & \textbf{1.00} & \textbf{1.00} & 1.83 & \textbf{1.00} \\
Pocket2mol & 1.05 & \textbf{1.00} & 1.90 & \textbf{1.00} \\
PocketFlow & \textbf{1.00} & \textbf{1.00} & \textbf{1.00} & 1.61 \\
ResGen & \textbf{1.00} & \textbf{1.00} & \textbf{1.00} & 1.16 \\
DST & 1.41 & 1.41 & 1.41 & 1.41 \\
graph ga & \textbf{1.00} & \textbf{1.00} & \textbf{1.00} & \textbf{1.00} \\
mimosa & 1.41 & 1.41 & 1.41 & 1.41 \\
moldqn & 1.51 & 1.65 & 1.51 & 1.62 \\
Pasithea & 1.41 & 1.41 & 1.41 & 1.41 \\
reinvent & 1.67 & 1.67 & 1.67 & 1.67 \\
screening & 1.51 & 1.51 & 1.51 & 1.51 \\
selfies vae bo & 1.75 & 1.75 & 1.75 & 1.75 \\
smiles ga & 1.61 & 1.61 & 1.60 & 1.61 \\
smiles lstm hc & 1.60 & 1.60 & 1.60 & 1.60 \\
smiles vae bo & 1.75 & 1.75 & 1.75 & 1.75 \\
\bottomrule
\end{tabular}
\end{sc}
\end{small}
\end{center}

\begin{center}
\begin{small}
\begin{sc}
\begin{tabular}{cccccccccccccc}
\toprule
Model \ & 4unn & 5mo4 & 7l11 \\
\midrule
3DSBDD & \textbf{1.00} & \textbf{1.00} & \textbf{1.00} \\
AutoGrow4 & 1.74 & \textbf{1.00} & 1.83 \\
Pocket2mol & \textbf{1.00} & \textbf{1.00} & \textbf{1.00} \\
PocketFlow & 1.11 & 1.54 & \textbf{1.00} \\
ResGen & \textbf{1.00} & \textbf{1.00} & \textbf{1.00} \\
DST & 1.41 & 1.41 & 1.41 \\
graph ga & \textbf{1.00} & \textbf{1.00} & \textbf{1.00} \\
mimosa & 1.41 & 1.41 & 1.41 \\
moldqn & 1.98 & 2.04 & 1.51 \\
Pasithea & 1.41 & 1.41 & 1.41 \\
reinvent & 1.67 & 1.67 & 1.67 \\
screening & 1.51 & 1.51 & 1.51 \\
selfies vae bo & 1.75 & 1.75 & 1.75 \\
smiles ga & 1.60 & 1.61 & 1.61 \\
smiles lstm hc & 1.60 & 1.60 & 1.60 \\
smiles vae bo & 1.75 & 1.75 & 1.75 \\
\bottomrule
\end{tabular}
\end{sc}
\end{small}
\end{center}
\end{table*}

\begin{table*}[h]
\caption{Top 10 SA score for each target protein. } 
\label{top10-sa-table}
\vskip 0.15in
\begin{center}
\begin{small}
\begin{sc}
\begin{tabular}{cccccccccccccc}
\toprule
Model \ & 1iep & 3eml & 3ny8 & 4rlu \\
\midrule
3DSBDD & 2.20 $\pm$ 0.38 & 1.54 $\pm$ 0.24 & 1.52 $\pm$ 0.34 & 2.80 $\pm$ 0.60 \\
AutoGrow4 & 1.04 $\pm$ 0.13 & 1.74 $\pm$ 0.26 & 1.91 $\pm$ 0.04 & 1.55 $\pm$ 0.31 \\
Pocket2mol & 1.36 $\pm$ 0.14 & \textbf{1.02} $\pm$ 0.03 & 2.07 $\pm$ 0.08 & \textbf{1.09} $\pm$ 0.08 \\
PocketFlow & \textbf{1.00} $\pm$ 0.00 & 1.46 $\pm$ 0.19 & 1.33 $\pm$ 0.18 & 1.61 $\pm$ 0.00 \\
ResGen & 1.03 $\pm$ 0.04 & 1.08 $\pm$ 0.08 & \textbf{1.07} $\pm$ 0.07 & 1.40 $\pm$ 0.13 \\
DST & 1.62 $\pm$ 0.09 & 1.62 $\pm$ 0.09 & 1.62 $\pm$ 0.09 & 1.62 $\pm$ 0.09 \\
graph ga & 1.37 $\pm$ 0.19 & 1.26 $\pm$ 0.17 & 1.32 $\pm$ 0.21 & 1.13 $\pm$ 0.14 \\
mimosa & 1.62 $\pm$ 0.09 & 1.62 $\pm$ 0.09 & 1.62 $\pm$ 0.09 & 1.62 $\pm$ 0.09 \\
moldqn & 2.53 $\pm$ 0.37 & 2.39 $\pm$ 0.36 & 2.61 $\pm$ 0.37 & 2.46 $\pm$ 0.32 \\
Pasithea & 1.67 $\pm$ 0.10 & 1.67 $\pm$ 0.10 & 1.67 $\pm$ 0.10 & 1.67 $\pm$ 0.10 \\
reinvent & 1.85 $\pm$ 0.08 & 1.88 $\pm$ 0.09 & 1.86 $\pm$ 0.09 & 1.88 $\pm$ 0.09 \\
screening & 1.63 $\pm$ 0.07 & 1.63 $\pm$ 0.07 & 1.63 $\pm$ 0.07 & 1.63 $\pm$ 0.07 \\
selfies vae bo & 1.90 $\pm$ 0.06 & 1.90 $\pm$ 0.06 & 1.90 $\pm$ 0.06 & 1.90 $\pm$ 0.06 \\
smiles ga & 1.90 $\pm$ 0.11 & 1.96 $\pm$ 0.14 & 1.86 $\pm$ 0.13 & 1.90 $\pm$ 0.11 \\
smiles lstm hc & 1.77 $\pm$ 0.09 & 1.77 $\pm$ 0.09 & 1.77 $\pm$ 0.09 & 1.77 $\pm$ 0.09 \\
smiles vae bo & 1.90 $\pm$ 0.06 & 1.90 $\pm$ 0.06 & 1.90 $\pm$ 0.06 & 1.90 $\pm$ 0.06 \\
\bottomrule
\end{tabular}
\end{sc}
\end{small}
\end{center}

\begin{center}
\begin{small}
\begin{sc}
\begin{tabular}{cccccccccccccc}
\toprule
Model \ & 4unn & 5mo4 & 7l11 \\
\midrule
3DSBDD & 2.20 $\pm$ 0.38 & 1.54 $\pm$ 0.24 & 1.52 $\pm$ 0.34 & 2.80 $\pm$ 0.60 \\
AutoGrow4 & 1.04 $\pm$ 0.13 & 1.74 $\pm$ 0.26 & 1.91 $\pm$ 0.04 & 1.55 $\pm$ 0.31 \\
Pocket2mol & 1.36 $\pm$ 0.14 & \textbf{1.02} $\pm$ 0.03 & 2.07 $\pm$ 0.08 & \textbf{1.09} $\pm$ 0.08 \\
PocketFlow & \textbf{1.00} $\pm$ 0.00 & 1.46 $\pm$ 0.19 & 1.33 $\pm$ 0.18 & 1.61 $\pm$ 0.00 \\
ResGen & 1.03 $\pm$ 0.04 & 1.08 $\pm$ 0.08 & \textbf{1.07} $\pm$ 0.07 & 1.40 $\pm$ 0.13 \\
DST & 1.62 $\pm$ 0.09 & 1.62 $\pm$ 0.09 & 1.62 $\pm$ 0.09 & 1.62 $\pm$ 0.09 \\
graph ga & 1.37 $\pm$ 0.19 & 1.26 $\pm$ 0.17 & 1.32 $\pm$ 0.21 & 1.13 $\pm$ 0.14 \\
mimosa & 1.62 $\pm$ 0.09 & 1.62 $\pm$ 0.09 & 1.62 $\pm$ 0.09 & 1.62 $\pm$ 0.09 \\
moldqn & 2.53 $\pm$ 0.37 & 2.39 $\pm$ 0.36 & 2.61 $\pm$ 0.37 & 2.46 $\pm$ 0.32 \\
Pasithea & 1.67 $\pm$ 0.10 & 1.67 $\pm$ 0.10 & 1.67 $\pm$ 0.10 & 1.67 $\pm$ 0.10 \\
reinvent & 1.85 $\pm$ 0.08 & 1.88 $\pm$ 0.09 & 1.86 $\pm$ 0.09 & 1.88 $\pm$ 0.09 \\
screening & 1.63 $\pm$ 0.07 & 1.63 $\pm$ 0.07 & 1.63 $\pm$ 0.07 & 1.63 $\pm$ 0.07 \\
selfies vae bo & 1.90 $\pm$ 0.06 & 1.90 $\pm$ 0.06 & 1.90 $\pm$ 0.06 & 1.90 $\pm$ 0.06 \\
smiles ga & 1.90 $\pm$ 0.11 & 1.96 $\pm$ 0.14 & 1.86 $\pm$ 0.13 & 1.90 $\pm$ 0.11 \\
smiles lstm hc & 1.77 $\pm$ 0.09 & 1.77 $\pm$ 0.09 & 1.77 $\pm$ 0.09 & 1.77 $\pm$ 0.09 \\
smiles vae bo & 1.90 $\pm$ 0.06 & 1.90 $\pm$ 0.06 & 1.90 $\pm$ 0.06 & 1.90 $\pm$ 0.06 \\
\bottomrule
\end{tabular}
\end{sc}
\end{small}
\end{center}
\end{table*}

\begin{table*}[h]
\caption{Top 50 SA score for each target protein. } 
\label{top50-sa-table}
\vskip 0.15in
\begin{center}
\begin{small}
\begin{sc}
\begin{tabular}{cccccccccccccc}
\toprule
Model \ & 1iep & 3eml & 3ny8 & 4rlu \\
\midrule
3DSBDD & 3.49 $\pm$ 0.78 & 2.28 $\pm$ 0.48 & 2.95 $\pm$ 1.00 & 4.06 $\pm$ 0.73 \\
AutoGrow4 & 1.53 $\pm$ 0.26 & 2.00 $\pm$ 0.20 & 2.07 $\pm$ 0.13 & 1.93 $\pm$ 0.25 \\
Pocket2mol & 1.68 $\pm$ 0.19 & \textbf{1.32} $\pm$ 0.19 & 2.30 $\pm$ 0.15 & \textbf{1.30} $\pm$ 0.13 \\
PocketFlow & \textbf{1.00} $\pm$ 0.00 & 1.88 $\pm$ 0.25 & 1.73 $\pm$ 0.28 & 1.67 $\pm$ 0.09 \\
ResGen & 1.36 $\pm$ 0.21 & 1.71 $\pm$ 0.41 & \textbf{1.39} $\pm$ 0.21 & 1.77 $\pm$ 0.24 \\
DST & 1.82 $\pm$ 0.13 & 1.82 $\pm$ 0.13 & 1.82 $\pm$ 0.13 & 1.82 $\pm$ 0.13 \\
graph ga & 1.80 $\pm$ 0.26 & 1.77 $\pm$ 0.30 & 1.83 $\pm$ 0.30 & 1.65 $\pm$ 0.32 \\
mimosa & 1.82 $\pm$ 0.13 & 1.82 $\pm$ 0.13 & 1.82 $\pm$ 0.13 & 1.82 $\pm$ 0.13 \\
moldqn & 3.11 $\pm$ 0.42 & 3.07 $\pm$ 0.45 & 3.18 $\pm$ 0.41 & 2.95 $\pm$ 0.33 \\
Pasithea & 1.85 $\pm$ 0.12 & 1.85 $\pm$ 0.12 & 1.85 $\pm$ 0.12 & 1.85 $\pm$ 0.12 \\
reinvent & 2.32 $\pm$ 0.35 & 2.37 $\pm$ 0.35 & 2.35 $\pm$ 0.36 & 2.38 $\pm$ 0.35 \\
screening & 1.83 $\pm$ 0.12 & 1.83 $\pm$ 0.12 & 1.83 $\pm$ 0.12 & 1.83 $\pm$ 0.12 \\
selfies vae bo & 2.15 $\pm$ 0.18 & 2.15 $\pm$ 0.18 & 2.15 $\pm$ 0.18 & 2.15 $\pm$ 0.18 \\
smiles ga & 2.27 $\pm$ 0.24 & 2.45 $\pm$ 0.30 & 2.24 $\pm$ 0.25 & 2.23 $\pm$ 0.22 \\
smiles lstm hc & 2.00 $\pm$ 0.14 & 2.00 $\pm$ 0.14 & 2.00 $\pm$ 0.14 & 2.00 $\pm$ 0.14 \\
smiles vae bo & 2.15 $\pm$ 0.18 & 2.15 $\pm$ 0.18 & 2.15 $\pm$ 0.18 & 2.15 $\pm$ 0.18 \\
\bottomrule
\end{tabular}
\end{sc}
\end{small}
\end{center}

\begin{center}
\begin{small}
\begin{sc}
\begin{tabular}{cccccccccccccc}
\toprule
Model \ & 4unn & 5mo4 & 7l11 \\
\midrule
3DSBDD & 2.14 $\pm$ 0.73 & 3.18 $\pm$ 0.87 & 2.52 $\pm$ 0.90 \\
AutoGrow4 & 2.00 $\pm$ 0.14 & 1.94 $\pm$ 0.22 & 2.06 $\pm$ 0.10 \\
Pocket2mol & \textbf{1.60} $\pm$ 0.17 & \textbf{1.26} $\pm$ 0.14 & 1.92 $\pm$ 0.26 \\
PocketFlow & 1.92 $\pm$ 0.27 & 1.72 $\pm$ 0.15 & \textbf{1.43} $\pm$ 0.20 \\
ResGen & 2.34 $\pm$ 0.46 & 1.88 $\pm$ 0.32 & 1.57 $\pm$ 0.31 \\
DST & 1.82 $\pm$ 0.13 & 1.82 $\pm$ 0.13 & 1.82 $\pm$ 0.13 \\
graph ga & 1.95 $\pm$ 0.37 & 1.76 $\pm$ 0.29 & 1.79 $\pm$ 0.35 \\
mimosa & 1.82 $\pm$ 0.13 & 1.82 $\pm$ 0.13 & 1.82 $\pm$ 0.13 \\
moldqn & 3.24 $\pm$ 0.38 & 3.16 $\pm$ 0.36 & 3.06 $\pm$ 0.39 \\
Pasithea & 1.85 $\pm$ 0.12 & 1.85 $\pm$ 0.12 & 1.85 $\pm$ 0.12 \\
reinvent & 2.37 $\pm$ 0.35 & 2.36 $\pm$ 0.34 & 2.36 $\pm$ 0.34 \\
screening & 1.83 $\pm$ 0.12 & 1.83 $\pm$ 0.12 & 1.83 $\pm$ 0.12 \\
selfies vae bo & 2.15 $\pm$ 0.18 & 2.15 $\pm$ 0.18 & 2.15 $\pm$ 0.18 \\
smiles ga & 2.23 $\pm$ 0.24 & 2.30 $\pm$ 0.23 & 2.39 $\pm$ 0.27 \\
smiles lstm hc & 2.00 $\pm$ 0.14 & 2.00 $\pm$ 0.14 & 2.00 $\pm$ 0.14 \\
smiles vae bo & 2.15 $\pm$ 0.18 & 2.15 $\pm$ 0.18 & 2.15 $\pm$ 0.18 \\
\bottomrule
\end{tabular}
\end{sc}
\end{small}
\end{center}
\end{table*}

\begin{table*}[h]
\caption{Top 100 SA score for each target protein. } 
\label{top100-sa-table}
\vskip 0.15in
\begin{center}
\begin{small}
\begin{sc}
\begin{tabular}{cccccccccccccc}
\toprule
Model \ & 1iep & 3eml & 3ny8 & 4rlu \\
\midrule
3DSBDD & 4.05 $\pm$ 0.79 & 2.85 $\pm$ 0.69 & 3.76 $\pm$ 1.09 & 4.46 $\pm$ 0.66 \\
AutoGrow4 & 1.72 $\pm$ 0.28 & 2.17 $\pm$ 0.23 & 2.29 $\pm$ 0.26 & 2.11 $\pm$ 0.25 \\
Pocket2mol & 1.86 $\pm$ 0.23 & \textbf{1.49} $\pm$ 0.22 & 2.50 $\pm$ 0.23 & \textbf{1.44} $\pm$ 0.17 \\
PocketFlow & \textbf{1.10} $\pm$ 0.16 & 2.11 $\pm$ 0.30 & 2.06 $\pm$ 0.40 & 1.87 $\pm$ 0.24 \\
ResGen & 1.64 $\pm$ 0.33 & 2.24 $\pm$ 0.65 & \textbf{1.66} $\pm$ 0.32 & 2.10 $\pm$ 0.38 \\
DST & 1.93 $\pm$ 0.14 & 1.93 $\pm$ 0.14 & 1.93 $\pm$ 0.14 & 1.93 $\pm$ 0.14 \\
graph ga & 2.03 $\pm$ 0.30 & 2.02 $\pm$ 0.33 & 2.09 $\pm$ 0.34 & 1.95 $\pm$ 0.38 \\
mimosa & 1.93 $\pm$ 0.14 & 1.93 $\pm$ 0.14 & 1.93 $\pm$ 0.14 & 1.93 $\pm$ 0.14 \\
moldqn & 3.53 $\pm$ 0.52 & 3.49 $\pm$ 0.53 & 3.59 $\pm$ 0.51 & 3.30 $\pm$ 0.44 \\
Pasithea & 1.97 $\pm$ 0.15 & 1.97 $\pm$ 0.15 & 1.97 $\pm$ 0.15 & 1.97 $\pm$ 0.15 \\
reinvent & 2.96 $\pm$ 0.80 & 3.03 $\pm$ 0.85 & 3.02 $\pm$ 0.86 & 3.03 $\pm$ 0.84 \\
screening & 1.94 $\pm$ 0.14 & 1.94 $\pm$ 0.14 & 1.94 $\pm$ 0.14 & 1.94 $\pm$ 0.14 \\
selfies vae bo & 2.42 $\pm$ 0.31 & 2.42 $\pm$ 0.31 & 2.42 $\pm$ 0.31 & 2.42 $\pm$ 0.31 \\
smiles ga & 2.52 $\pm$ 0.31 & 2.76 $\pm$ 0.38 & 2.54 $\pm$ 0.35 & 2.51 $\pm$ 0.33 \\
smiles lstm hc & 2.14 $\pm$ 0.18 & 2.14 $\pm$ 0.18 & 2.14 $\pm$ 0.18 & 2.14 $\pm$ 0.18 \\
smiles vae bo & 2.42 $\pm$ 0.31 & 2.42 $\pm$ 0.31 & 2.42 $\pm$ 0.31 & 2.42 $\pm$ 0.31 \\

\bottomrule
\end{tabular}
\end{sc}
\end{small}
\end{center}

\begin{center}
\begin{small}
\begin{sc}
\begin{tabular}{cccccccccccccc}
\toprule
Model \ & 4unn & 5mo4 & 7l11 \\
\midrule
3DSBDD & 3.25 $\pm$ 1.29 & 4.28 $\pm$ 1.37 & 3.49 $\pm$ 1.17 \\
AutoGrow4 & 2.12 $\pm$ 0.16 & 2.11 $\pm$ 0.23 & 2.19 $\pm$ 0.16 \\
Pocket2mol & \textbf{1.76} $\pm$ 0.20 & \textbf{1.40} $\pm$ 0.18 & 2.17 $\pm$ 0.31 \\
PocketFlow & 2.20 $\pm$ 0.35 & 2.04 $\pm$ 0.35 & \textbf{1.67} $\pm$ 0.28 \\
ResGen & 2.70 $\pm$ 0.49 & 2.20 $\pm$ 0.40 & 1.97 $\pm$ 0.47 \\
DST & 1.93 $\pm$ 0.14 & 1.93 $\pm$ 0.14 & 1.93 $\pm$ 0.14 \\
graph ga & 2.22 $\pm$ 0.38 & 1.99 $\pm$ 0.32 & 2.05 $\pm$ 0.36 \\
mimosa & 1.93 $\pm$ 0.14 & 1.93 $\pm$ 0.14 & 1.93 $\pm$ 0.14 \\
moldqn & 3.60 $\pm$ 0.45 & 3.56 $\pm$ 0.48 & 3.46 $\pm$ 0.49 \\
Pasithea & 1.97 $\pm$ 0.15 & 1.97 $\pm$ 0.15 & 1.97 $\pm$ 0.15 \\
reinvent & 3.03 $\pm$ 0.85 & 3.01 $\pm$ 0.84 & 3.01 $\pm$ 0.84 \\
screening & 1.94 $\pm$ 0.14 & 1.94 $\pm$ 0.14 & 1.94 $\pm$ 0.14 \\
selfies vae bo & 2.42 $\pm$ 0.31 & 2.42 $\pm$ 0.31 & 2.42 $\pm$ 0.31 \\
smiles ga & 2.49 $\pm$ 0.32 & 2.54 $\pm$ 0.30 & 2.68 $\pm$ 0.36 \\
smiles lstm hc & 2.14 $\pm$ 0.18 & 2.14 $\pm$ 0.18 & 2.14 $\pm$ 0.18 \\
smiles vae bo & 2.42 $\pm$ 0.31 & 2.42 $\pm$ 0.31 & 2.42 $\pm$ 0.31 \\
\bottomrule
\end{tabular}
\end{sc}
\end{small}
\end{center}
\end{table*}

\begin{table*}[h]
\caption{Model Rankings based on QED}
\begin{center}
\begin{small}
\begin{sc}
\begin{tabular}{ccccc}
\toprule
\textbf{model } & \textbf{  Top 1 Rank } & \textbf{  Top 10 Rank } & \textbf{  Top 100 Rank } & \textbf{  Overall Rank} \\
\midrule
3DSBDD         &          13 &           14 &            15 &            14 \\
AutoGrow4      &          14 &           13 &            12 &            13 \\
Pocket2mol     &          11 &           11 &             8 &            11 \\
PocketFlow     &          15 &           15 &            14 &            15 \\
ResGen         &          12 &           12 &            11 &            12 \\
DST            &           2 &            2 &             2 &             1 \\
graph ga       &           6 &            5 &             6 &             5 \\
mimosa         &           2 &            2 &             2 &             1 \\
moldqn         &          16 &           16 &            16 &            16 \\
Pasithea       &           2 &            4 &             4 &             4 \\
reinvent       &           4 &           10 &            13 &            10 \\
screening      &           5 &            1 &             1 &             3 \\
selfies vae bo &           8 &            8 &             9 &             8 \\
smiles ga      &          10 &            6 &             5 &             6 \\
smiles lstm hc &           7 &            7 &             7 &             6 \\
smiles vae bo  &           8 &            8 &             9 &             8 \\
\bottomrule

\label{tab:model_rankings_qed}
\end{tabular}
\end{sc}
\end{small}
\end{center}
\end{table*}

\begin{table*}[h]
\caption{Model Rankings based on SA}
\begin{center}
\begin{small}
\begin{sc}
\begin{tabular}{ccccc}
\toprule
\textbf{model } & \textbf{  Top 1 Rank } & \textbf{  Top 10 Rank } & \textbf{  Top 100 Rank } & \textbf{  Overall Rank} \\
\midrule
3DSBDD         &          12 &            6 &             1 &             6 \\
AutoGrow4      &          11 &            9 &             8 &             8 \\
Pocket2mol     &          14 &           14 &            16 &            16 \\
PocketFlow     &          13 &           13 &            15 &            14 \\
ResGen         &          15 &           16 &             9 &            13 \\
DST            &           9 &           11 &            13 &            11 \\
graph ga       &          16 &           15 &            10 &            14 \\
mimosa         &           9 &           11 &            14 &            12 \\
moldqn         &           3 &            1 &             2 &             1 \\
Pasithea       &           9 &            8 &            11 &             8 \\
reinvent       &           4 &            5 &             3 &             5 \\
screening      &           7 &           10 &            12 &            10 \\
selfies vae bo &           1 &            3 &             5 &             2 \\
smiles ga      &           5 &            2 &             4 &             4 \\
smiles lstm hc &           6 &            7 &             7 &             7 \\
smiles vae bo  &           1 &            3 &             5 &             2 \\
\bottomrule

\label{tab:model_rankings_sa}
\end{tabular}
\end{sc}
\end{small}
\end{center}
\end{table*}

\begin{table*}[h]
\caption{Model Rankings based on docking score}
\begin{center}
\begin{small}
\begin{sc}
\begin{tabular}{ccccc}
\toprule
\textbf{model } & \textbf{  Top 1 Rank } & \textbf{  Top 10 Rank } & \textbf{  Top 100 Rank } & \textbf{  Overall Rank} \\
\midrule
3DSBDD         &          12 &           12 &            13 &            12 \\
AutoGrow4      &           2 &            1 &             1 &             1 \\
Pocket2mol     &           1 &            2 &             2 &             2 \\
PocketFlow     &          11 &           14 &            14 &            13 \\
ResGen         &           3 &            3 &             7 &             4 \\
DST            &           6 &            5 &             4 &             5 \\
graph ga       &           9 &            9 &             9 &             9 \\
mimosa         &           5 &            6 &             5 &             6 \\
moldqn         &          16 &           16 &            16 &            16 \\
Pasithea       &           7 &            7 &             6 &             7 \\
reinvent       &          15 &           15 &            15 &            15 \\
screening      &           4 &            4 &             3 &             3 \\
selfies vae bo &          10 &           10 &            11 &            10 \\
smiles ga      &          14 &           13 &            12 &            13 \\
smiles lstm hc &           8 &            8 &             8 &             8 \\
smiles vae bo  &          13 &           11 &            10 &            11 \\
\bottomrule

\label{tab:model_rankings_docking}
\end{tabular}
\end{sc}
\end{small}
\end{center}
\end{table*}

\begin{table*}[h]
\caption{Model Rankings based on average Molecule Generation Metrics across all target proteins} 
\label{tab:model_rankings_metrics}
\vskip 0.15in
\begin{center}
\begin{small}
\begin{sc}
\begin{tabular}{ccccccc}
\toprule
Model \ & diversity & validity & uniqueness \\
\midrule
3DSBDD &  0.83 & 0.63 & 0.63 \\
AutoGrow4 &0.84 & 1.00 & 0.29 \\
Pocket2mol & 0.86 & 1.00 & 1.00 \\
PocketFlow & 0.90 & 1.00 & 0.87 \\
ResGen & 0.83 & 1.00 & 1.00 \\
DST & 0.88 & 1.00 & 1.00  \\
graph ga &  0.91 & 1.00 & 1.00 \\
mimosa & 0.88 & 1.00 &  1.00 \\
moldqn & 0.91 & 1.00 & 1.00 \\
Pasithea & 0.89 &  1.00 &  1.00 \\
reinvent & 0.88 & 1.00 & 1.00 \\
screening & 0.88 & 1.00 & 1.00 \\
selfies vae bo &  0.88 & 1.00 & 1.00\\
smiles ga & 0.88 & 1.00 & 1.00 \\
smiles lstm hc & 0.89 & 1.00 &  1.00 \\
smiles vae bo &  0.88 & 1.00 &  1.00 \\

\bottomrule
\end{tabular}
\end{sc}
\end{small}
\end{center}

\begin{center}
\begin{small}
\begin{sc}
\begin{tabular}{ccccccc}
\toprule
Model \ & Diversity Rank & validity Rank & uniqueness Rank \\
\midrule
3DSBDD & 14.00 & 2.00 &  3.00 \\
AutoGrow4 & 13.00 &1.00 & 4.00 \\
Pocket2mol & 12.00 &1.00 & 1.00 \\
PocketFlow & 3.00 & 1.00 & 2.00 \\
ResGen &   15.00 & 1.00 &  1.00 \\
DST & 7.00 & 1.00 & 1.00 \\
graph ga & 2.00 & 1.00 & 1.00 \\
mimosa & 8.00 & 1.00 & 1.00 \\
moldqn & 1.00 &  1.00 & 1.00 \\
Pasithea & 4.00 &  1.00 &  1.00 \\
reinvent & 10.00 & 1.00 & 1.00 \\
screening & 9.00 & 1.00 & 1.00 \\
selfies vae bo & 6.00 & 1.00 & 1.00 \\
smiles ga & 11.00 & 1.00 & 1.00 \\
smiles lstm hc & 5.00 & 1.00 & 1.00 \\
smiles vae bo & 6.00 & 1.00 & 1.00 \\
\bottomrule
\end{tabular}
\end{sc}
\end{small}
\end{center}
\end{table*}

\begin{table}[t]
\caption{Number of molecules generated under given 96 hours. }
\label{avg_gen_table}
\vskip 0.15in
\begin{center}
\begin{small}
\begin{sc}
\begin{tabular}{cccccccc}
\toprule
Model & 1iep & 3eml & 3ny8 & 4rlu & 4unn & 5mo4 & 7l11 \\
\midrule
3DSBDD & 1002 & 715 & 753 & 826 & 900 & 616 & 589 \\

AutoGrow4 & 1429 & 1438 & 1319 & 1272 & 1552 & 1496 & 1301 \\

Pocket2mol & 1038 & 1020 & 900 & 900 & 841 & 900 & 900 \\

PocketFlow & 1000 & 1000 & 1000 & 1000 & 1000 & 1000 & 1000 \\

ResGen & 800 & 800 & 800 & 800 & 322 & 369 & 527 \\

DST & 1001 & 1001 & 1001 & 1001 & 1001 & 1001 & 1001 \\

graph GA & 700 & 1001 & 700 & 700 & 300 & 1001 & 1001 \\

MIMOSA & 1001 & 1001 & 1001 & 1001 & 1001 & 1001 & 1001 \\

MoLDQN & 501 & 501 & 501 & 501 & 501 & 501 & 501 \\

Pasithea & 800 & 1000 & 800 & 800 & 1000 & 1000 & 1000 \\

reinvent & 100 & 100 & 100 & 100 & 100 & 100 & 100 \\

screening & 1000 & 1000 & 1000 & 1000 & 1000 & 1000 & 1000 \\

selfies vae bo & 200 & 200 & 200 & 200 & 200 & 200 & 200 \\

smiles ga & 525 & 441 & 615 & 618 & 808 & 710 & 376 \\

smiles lstm hc & 501 & 501 & 501 & 501 & 501 & 501 & 501 \\

smiles vae bo & 200 & 200 & 200 & 200 & 200 & 200 & 200 \\
\bottomrule
\end{tabular}
\end{sc}
\end{small}
\end{center}
\end{table}

\begin{figure}
  \centering
  \begin{subfigure}[b]{0.45\textwidth}
    \centering
    \includegraphics[width=\textwidth]{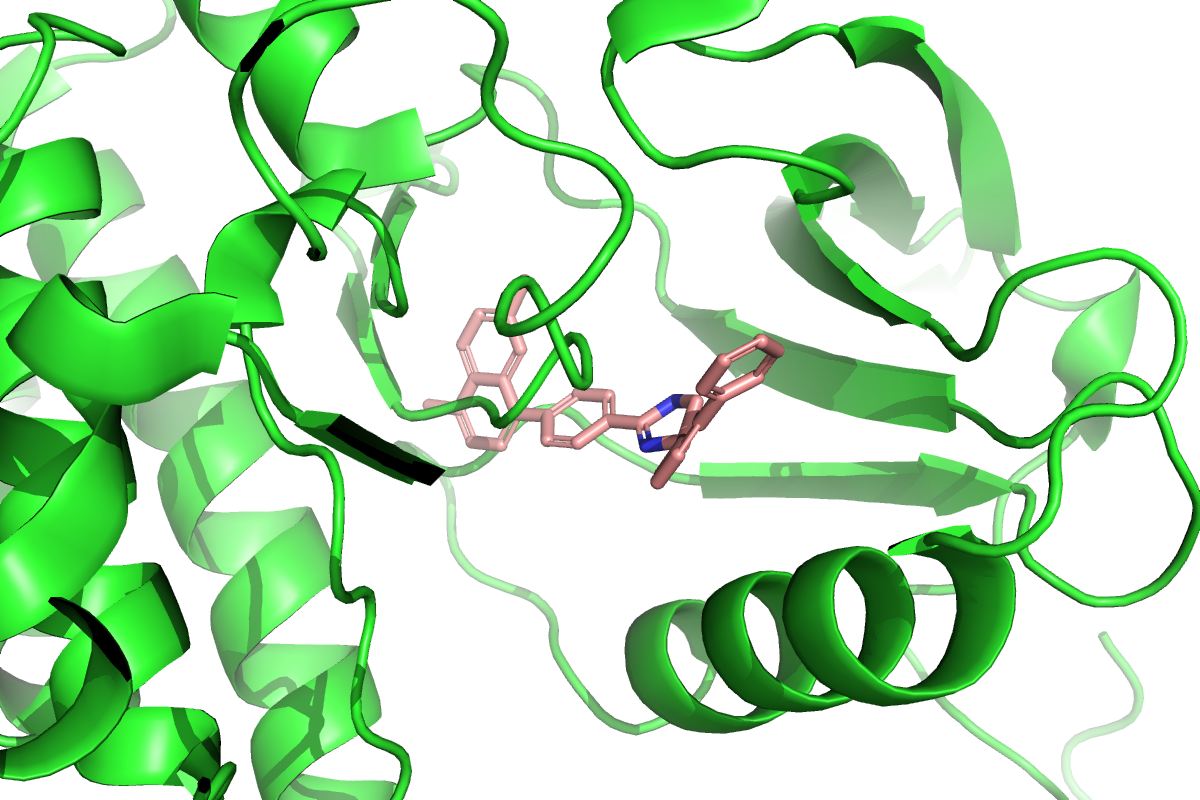}
    \caption{Model: PocketFlow, PDB: 1iep, -13.9 kcal/mol}
    \label{fig:subfig1}
  \end{subfigure}
  \hfill
  \begin{subfigure}[b]{0.45\textwidth}
    \centering
    \includegraphics[width=\textwidth]{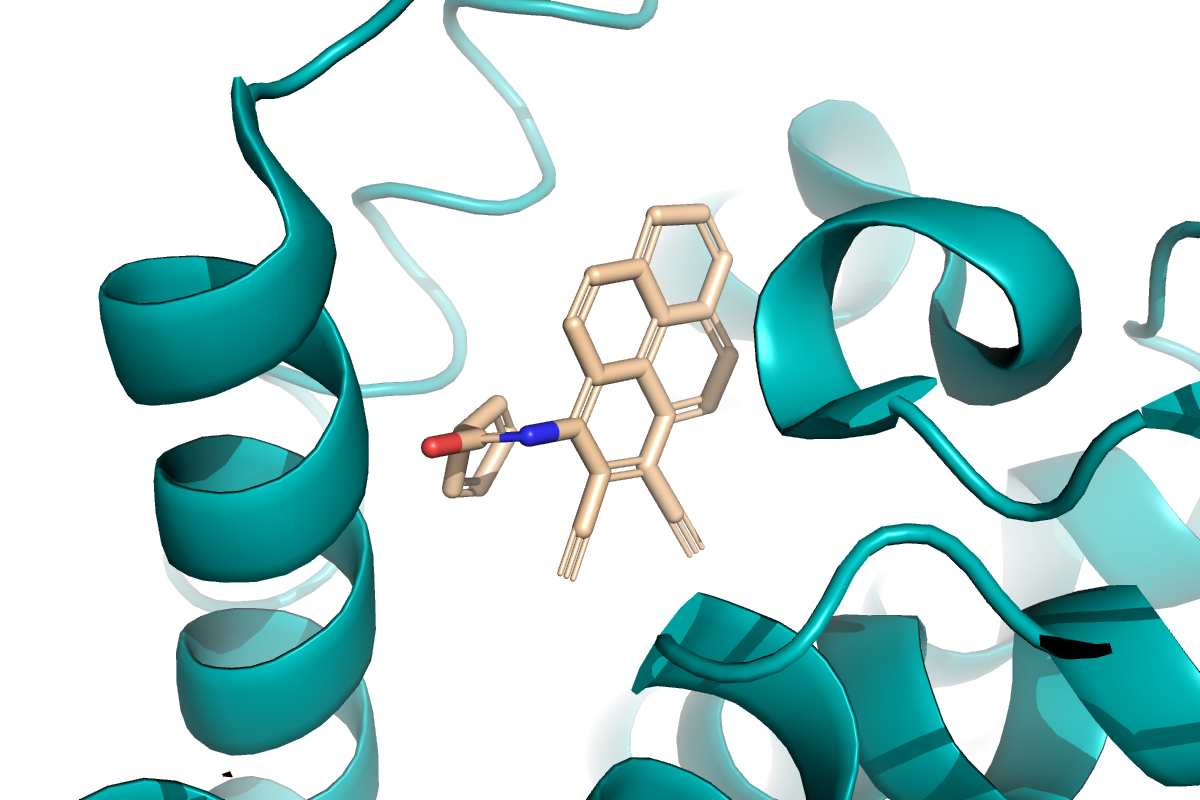}
    \caption{Model: AutoGrow4, PDB: 3eml, -13.3 kcal/mol}
    \label{fig:subfig2}
  \end{subfigure}
  
  \begin{subfigure}[b]{0.45\textwidth}
    \centering
    \includegraphics[width=\textwidth]{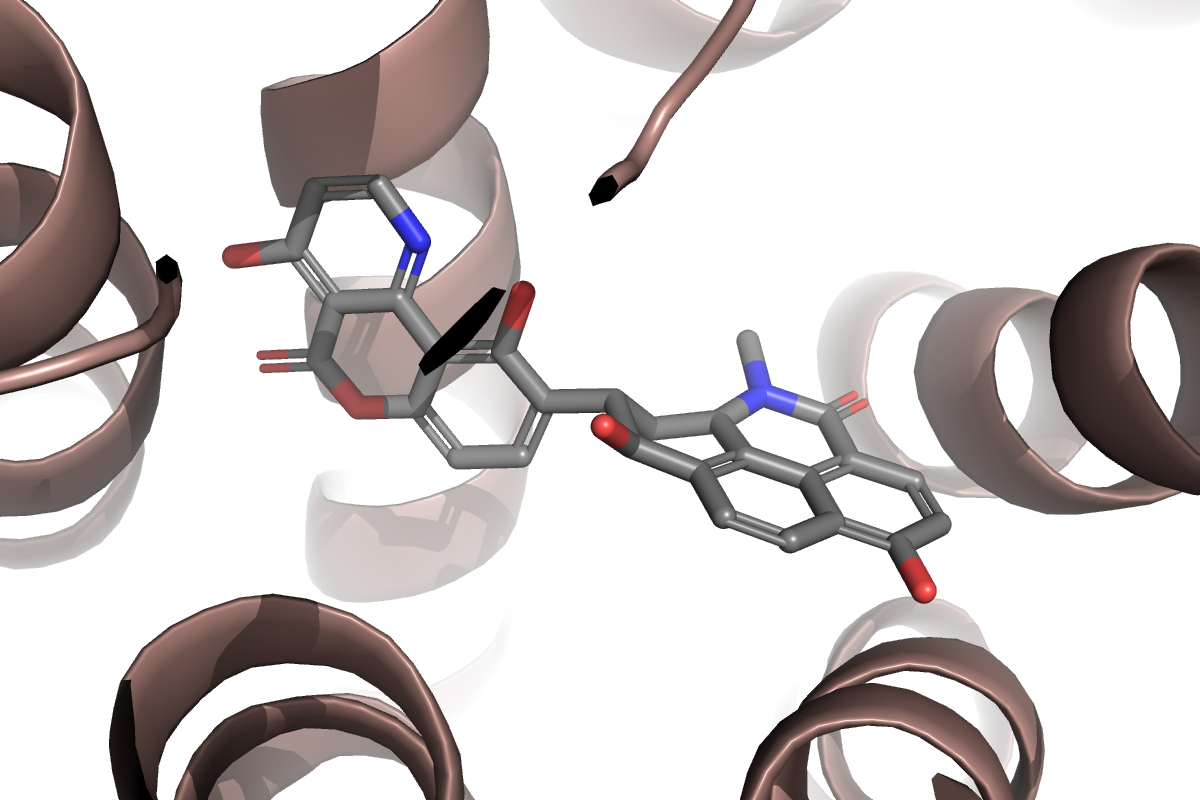}
    \caption{Model: Pocket2mol, PDB: 3ny8, -12.2 kcal/mol}
    \label{fig:subfig3}
  \end{subfigure}
  \hfill
  \begin{subfigure}[b]{0.45\textwidth}
    \centering
    \includegraphics[width=\textwidth]{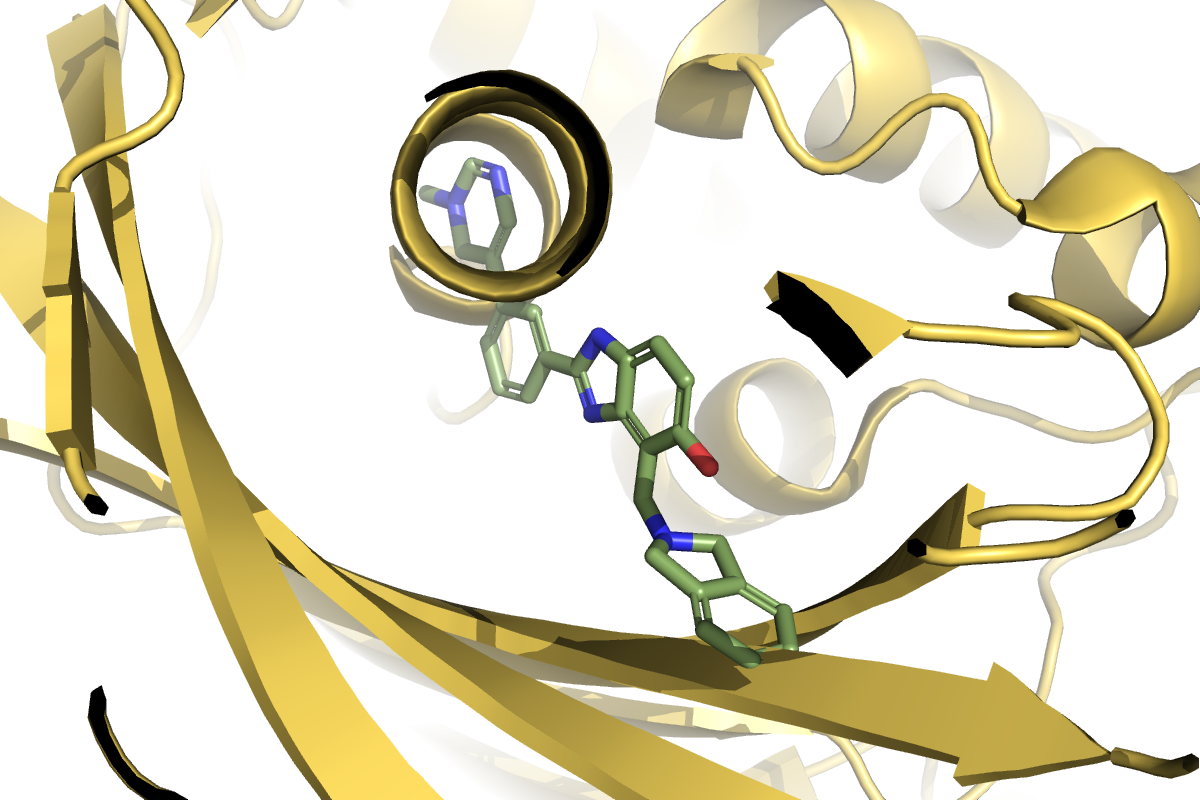}
    \caption{Model: ResGen, PDB: 4rlu, -12.6 kcal/mol}
    \label{fig:subfig4}
  \end{subfigure}
  
  \begin{subfigure}[b]{0.45\textwidth}
    \centering
    \includegraphics[width=\textwidth]{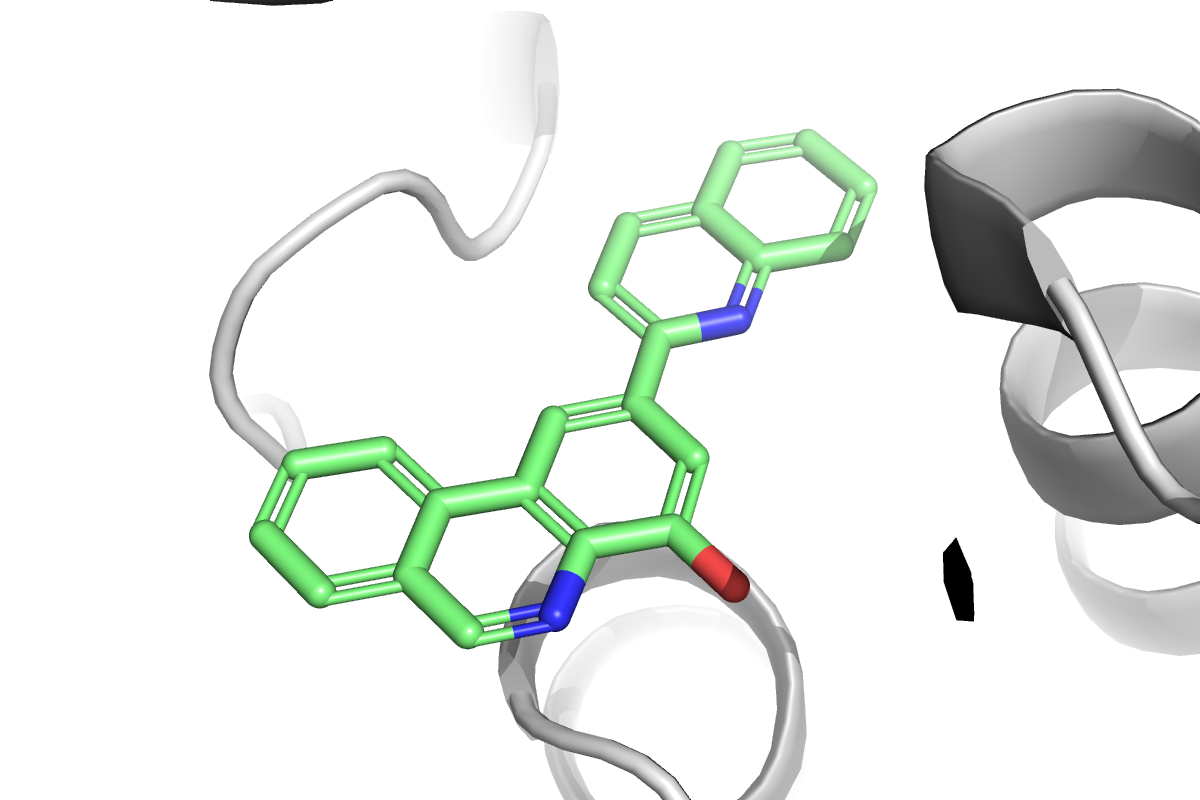}
    \caption{Model: Pocket2mol, PDB: 5mo4, -11.9 kcal/mol}
    \label{fig:subfig5}
  \end{subfigure}
  \hfill
  \begin{subfigure}[b]{0.45\textwidth}
    \centering
    \includegraphics[width=\textwidth]{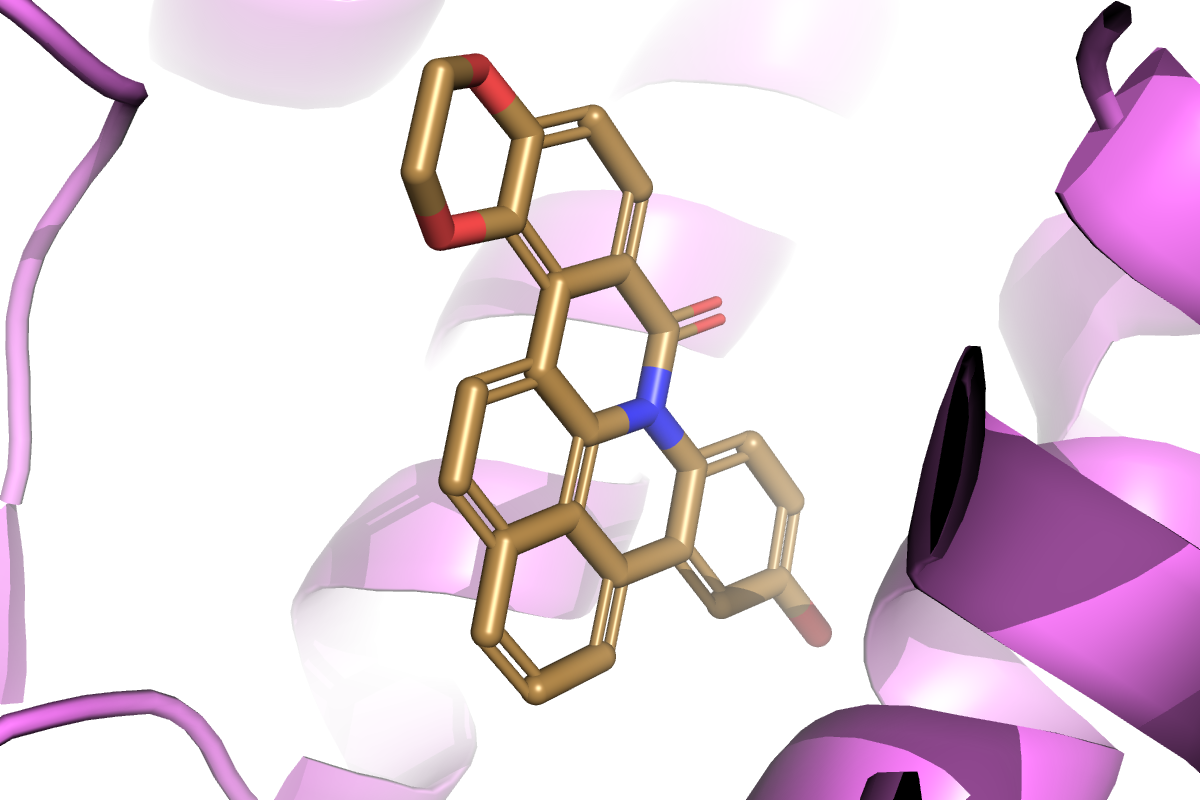}
    \caption{Model: Pocket2mol, PDB: 4unn, -12.8 kcal/mol}
    \label{fig:subfig6}
  \end{subfigure}
  
  \begin{subfigure}[b]{0.45\textwidth}
    \centering
    \includegraphics[width=\textwidth]{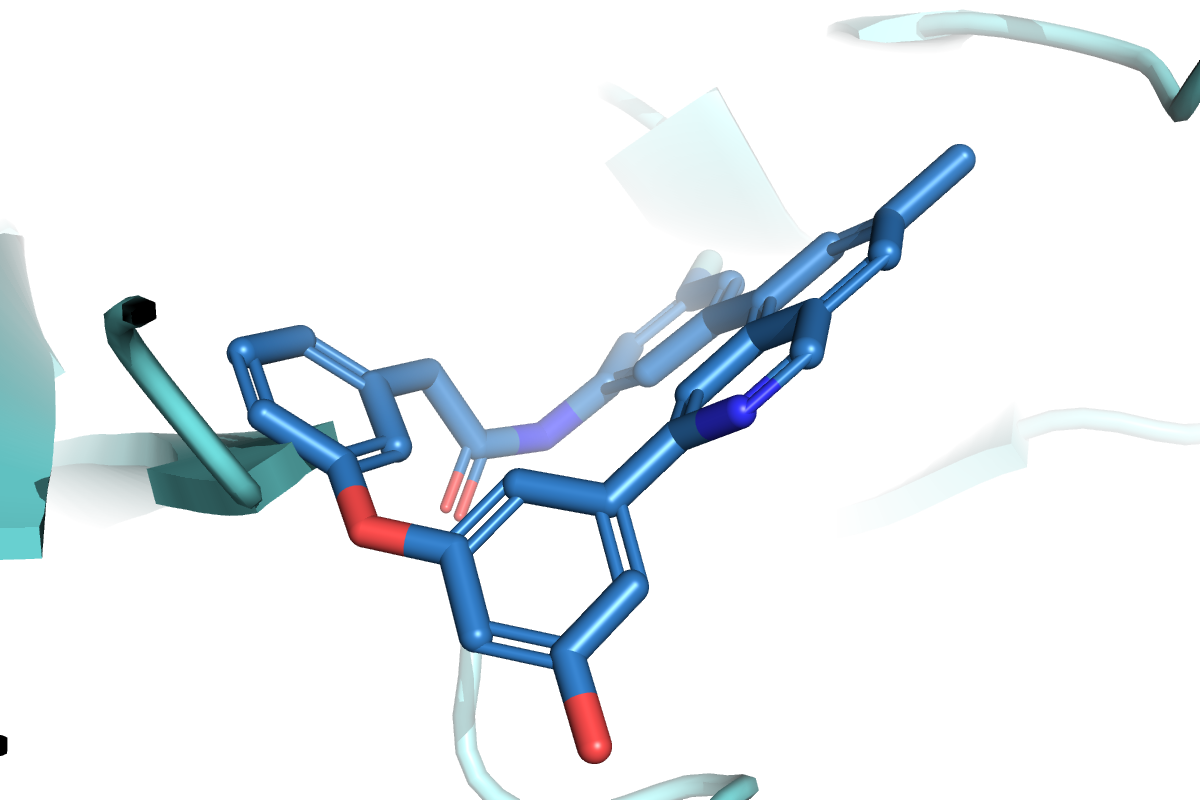}
    \caption{Model: Pocket2mol, PDB: 7l11, -10.4 kcal/mol}
    \label{fig:subfig7}
  \end{subfigure}
  
  \caption{Examples of best generated molecules for each PDB}
  \label{fig:pose}
\end{figure}

\end{document}